\documentclass[reprint,onecolumn,showpacs,preprintnumbers,amsmath, aps,pre,amssymb,notitlepage]{revtex4-1}

\usepackage{comment}
\usepackage{wrapfig,psfig,epsf,epsfig,subfigure}
\usepackage{graphicx}

\usepackage[ruled,vlined,commentsnumbered]{algorithm2e}
\usepackage{paralist}
\usepackage{lipsum}
\usepackage{bbm}
\usepackage{times}

\usepackage[breaklinks]{hyperref}

\hypersetup{
    bookmarks=true,         
    unicode=false,          
    pdftoolbar=true,        
    pdfmenubar=true,        
    pdffitwindow=false,     
    pdftitle={Scalable Models for Computing Hierarchies in Information Networks},    
    pdfauthor={Baoxu Shi and Tim Weninger},     
    pdfnewwindow=true,      
    colorlinks=true,       
    linkcolor=blue,          
    citecolor=blue,        
    filecolor=blue,      
    urlcolor=blue           
}

\newcommand{\tg}[1]{}

\usepackage{expl3}
\ExplSyntaxOn
\newcommand\latinabbrev[1]{
  \peek_meaning:NTF . {
    #1\@}%
  { \peek_catcode:NTF a {
      #1.\@ }%
    {#1.\@}}}
\ExplSyntaxOff

\newcommand{\ignore}[1]{}

\newcommand\ie{\emph{i.e.}}

\newcommand\bcmdtab{\noindent\bgroup\tabcolsep=0pt%
  \begin{tabular}{@{}p{10pc}@{}p{20pc}@{}}}
\newcommand\ecmdtab{\end{tabular}\egroup}

 \newcommand{\remove}[1]{}
\usepackage{color}

\newcommand{\nop}[1]{}

\usepackage{color}

\usepackage[nolist]{acronym}

\begin{document}

\label{firstpage}

\title{Scalable Models for Computing Hierarchies in Information Networks}

\author{Baoxu Shi}
 \email{bshi@nd.edu}
\author{Tim Weninger}
 \email{tweninge@nd.edu}
\affiliation{%
$^{\ast\dag}$Department of Computer Science and Engineering,
University of Notre Dame, Notre Dame, Indiana, USA
}%
\date{\today}

\begin{abstract}
Information hierarchies are organizational structures that often used to organize and present large and complex information as well as provide a mechanism for effective human navigation. Fortunately, many statistical and computational models exist that automatically generate hierarchies; however, the existing approaches do not consider linkages in information {\em networks} that are increasingly common in real-world scenarios. Current approaches also tend to present topics as an abstract probably distribution over words, etc rather than as tangible nodes from the original network. Furthermore, the statistical techniques present in many previous works are not yet capable of processing data at Web-scale. In this paper we present the Hierarchical Document Topic Model (HDTM), which uses a distributed vertex-programming process to calculate a nonparametric Bayesian generative model. Experiments on three medium size data sets and the entire Wikipedia dataset show that HDTM can infer accurate hierarchies even over large information networks.
\end{abstract}

\maketitle

\section{Introduction}
\label{sec:intro}

As the number of online resources and Web documents continues to increase, the need for better organizational structures that guide readers towards the information they seek increases. Hierarchies and taxonomies are invaluable tools for this purpose. Taxonomies are widely used in libraries via the Library of Congress System or the Dewey Decimal System, and hierarchies were a fixture of the early World Wide Web; perhaps the most famous example being the Yahoo search engine, which originally was a taxonomic-collection of hyperlinks organized by topic. These systems were developed because their effectiveness at topical organization and their logarithmic depth allowed users to quickly find the relevant documents for which they were searching. 

Unfortunately, taxonomy curation of documents, articles, books, etc. is mostly a manual process, which is only possible when the number of curated documents is relatively small. This process becomes increasingly impractical as the number of documents grows to Web-scale, and has motivated research towards the automatic inference of taxonomies~\citep{Adams2010,Blei2003,Ho2012,Chambers2010,Chang2009b,Mimno2007}. 

Most document repositories contain linkages between the documents creating a {\em document-graph}. These links provide, among other things, proper context to the terms and topics in each document. Document-graphs are especially common in nonfiction and scientific literature, where citations are viewed as inter-document links. Similarly, the World Wide Web (WWW) can be considered to be a single, very large document-graph, where Web pages represent documents and hyperlinks link documents. Web sites on the WWW could also be considered document graphs because a Web site is simply a subgraph of the WWW. Web site subgraphs, in particular, are a collection of documents with a specific and purposeful organizational structure, which are often carefully designed to guide the user from the entry page, {\em i.e.}, a homepage, to progressively more specific Web pages. 

Similarly, scientific literature can be categorized into a hierarchy of increasingly specific scientific topics by their citation links, and encyclopedia articles can be categorized into a hierarchy of increasingly specific articles by their cross references. Thus, we assert that most document-graphs, or, more generally, information networks, contain hidden node hierarchies. 

\subsection{Taxonomies versus Hierarchies}
In this paper we draw specific distinctions between a {\em hierarchy} and a {\em taxonomy}. A taxonomy is defined to be a classification of objects into increasingly finer granularities, where each non-leaf node is a conceptual combination of its children. A biological taxonomy is a canonical example of this definition because a classified species, say \textsf{homo sapiens} ({\em i.e.}, humans), can only be placed at a leaf in the taxonomy; the inner nodes, {\em e.g.}, \textsf{primate}, \textsf{mammal}, \textsf{animal}, do not declare new species, rather they are conceptual agglomerations of species. Furthermore, each species is described by its path through the taxonomy. For example, \textsf{homo sapien}s, can be more generally described as \textsf{primate}s, \textsf{mammal}s and as \textsf{animal}s (among others). 

A hierarchy, on the other hand, is an arrangement of objects where some objects are considered to be {\em above}, {\em below} or {\em at the same level as} others. This necessarily means that objects of a hierarchy live at the internal nodes. For example, a business, government or military chain of command is a hierarchy because the \textsf{president} is a specific object that is above the \textsf{general}s, which are above the \textsf{captain}s, and so on. In this case, high-level nodes like \textsf{president} are not agglomerations of \textsf{captain}s just as the \textsf{CEO} is not an aggregation of \textsf{manager}s or \textsf{clerk}s.

\subsection{Hierarchies of Documents}
\label{sec:inh_taxonomies}
In the similar studies, ``document hierarchies'' were not actually hierarchies of documents in the literal sense. For example, Hierarchical LDA (hLDA)~\citep{Blei2004, Blei2010}, TopicBlock~\citep{Ho2012}, and the Tree Structured Stick Breaking (TSSB) model~\citep{Adams2010} learn a conceptual taxonomy in which the non-leaf topics are a combination of words and does not represent a real document in the corpus; the hierarchical Pachinko allocation model (hPAM)~\citep{Mimno2007} constructs a tree-like conceptual taxonomy like hLDA, but where each topic can have multiple parents. 

In these related models, only the leaves contained the actual, literal documents. Contrary to our perspective, the internal nodes of existing models contain ephemeral word-topic distributions, rather than actual documents. See Figure~\ref{fig:inh_relatedwork} for a brief comparison of model outputs. The HDTM model introduced in this paper requires that inner nodes, which in previous work are made of ephemeral distributions, be literal documents. This requirement asserts that {\em some documents are more general than others}. Here we explore this assertion through examples and a review of similar assertions made in previous research.

\subsubsection{Web sites as Document Hierarchies}

A Web site $G$ can be viewed as a directed graph with Web pages as vertices $V$ and hyperlinks as directed edges $E$ between Web pages $v_x \rightarrow v_y$ -- excluding inter-site hyperlinks. In most cases, designating the Web site entry page as the root $r$ allows for a Web site to be viewed as a rooted directed graph. Web site creators and curators purposefully organize the hyperlinks between documents in a topically meaningful manner. As a result, Web documents further away from the root document typically contain more specific topics than Web documents graphically close to the root document. 

\begin{figure}
	\centering
		\includegraphics[width=\textwidth]{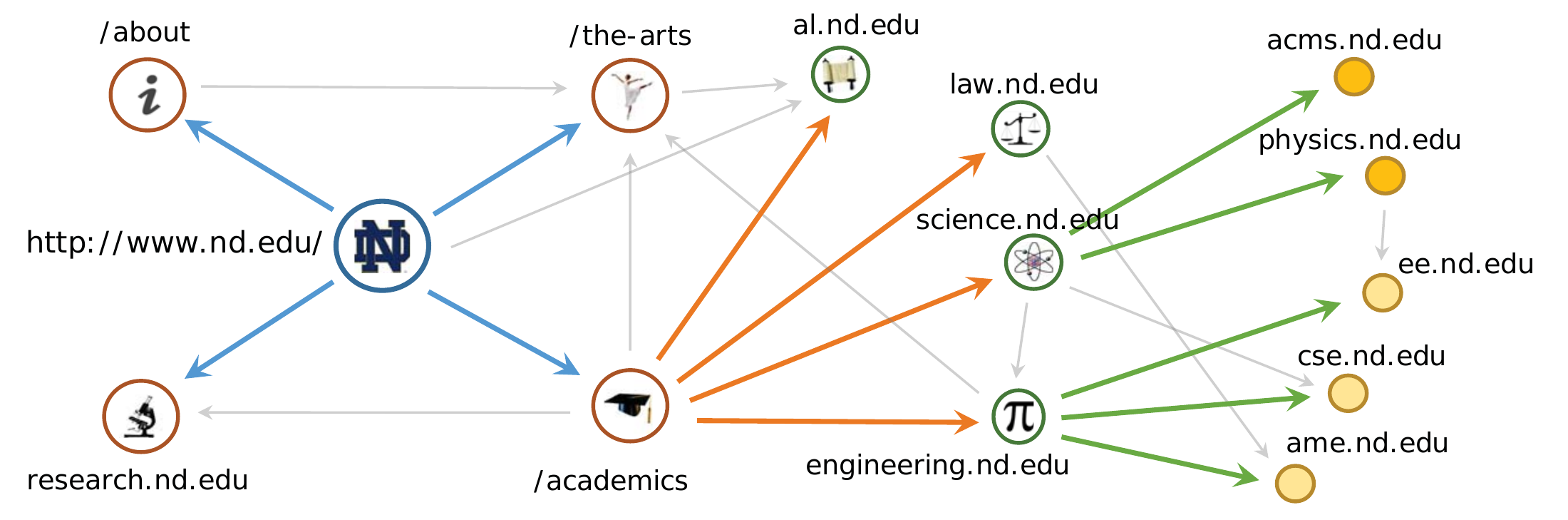}		
	\caption{Truncated Web site graph of the University of Notre Dame. Node color and size represent more-general to less-general nodes as determined by some hierarchy inference algorithm. Grey edges are hyperlinks that were not selected to be present in the inferred hierarchy.}
	\label{fig:inh_webtree}
\end{figure}

For example, the Web site at the University of Notre Dame, shown in Figure~\ref{fig:inh_webtree} contains a root Web document (the entry page), and dozens of children Web documents. Even with a very small subset of documents and edges, the corresponding Web graph can be quite complicated and messy. A breadth first traversal of the Web graph starting with the root node is a simple way to distill a document hierarchy from the Web graph. Unfortunately, a fixed breadth-first hierarchy cannot account for many of the intricacies of real world Web graphs.

For explanation purposes, let's assume there are four types of hyperlink edges in a Web site: (1) parent-to-child links, (2) upward links, (3) shortcuts, and (4) cross-topic links. Parent-to-child links direct the user from one Web page to a more topically specific Web page; {\em e.g.}, a hyperlink from {\small\url{engineering.nd.edu}} to {\small\url{cse.nd.edu}} is a parent-to-child hyperlink because computer science is topically more specific than engineering. Upward links are hyperlinks that reference a more general document; {\em e.g.}, there may exist a hyperlink from {\small\url{engineering.nd.edu}} to {\small\url{/the-arts}} because the engineering college would like to reference artistic happenings at the university. Shortcut links are hyperlinks that skip from very general Web documents to very specific Web documents as a way of featuring some specific topic; {\em e.g.}, if a computer science professor wins a prestigious award or grant, the professor's Web page may be linked to from the news section of the root Web page. Cross topic links are hyperlinks that move across topical subtrees; {\em e.g.}, the college of science may reference some working relationship with the college of engineering by creating a hyperlink between the two Web pages. 

Because the goal is to infer the document hierarchy, the HDTM model is, in a sense, trying to find parent-to-child links. In the event that there is more than one parent-to-child link to a particular Web page, the goal is to find the best topically-relevant parent for each Web document in the inferred hierarchy (except the root).

Web researchers and practitioners have used hyperlink structures to organize Web documents for many years. The PageRank and HITS algorithms are two of the most famous examples of information propagation through links. PageRank, for instance, uses the model of a random Web surfer ({\em i.e.} random walker), who randomly follows hyperlinks over the Web; a current measure of a Web page's authority corresponds to the probability that a random surfer lands upon that Web page -- the PageRank score. In the HDTM model, PageRank's notion of authority loosely corresponds to topical generality, that is, Web pages with a high random surfer probability are likely to be topically more general than others.

\subsubsection{Term propagation in Document Graphs}
The document-graph structure is also used to enrich the document description by adding features to improve retrieval performance. Some of the intuition behind these previous works are helpful in framing the generative model.

A limitation of the random walker model is that it only looks at the graphical structure of the network. The word distributions found in each document are clearly an important factor to consider when generating document hierarchies. Previous work by Song, {\em et al.}~\citep{Song2004} and Qin, {\em et al.}~\citep{Qin2005} show that a given Web page can be enriched by propagating information from its children. Their relevance propagation model modifies the language distribution of a Web page to be a mixture of itself and its children according to the formula: 

\begin{equation}
    f^{\prime}(w;d) = (1+\alpha)f(w;d) + \frac{(1-\alpha)}{|Child(d)|}\sum_{c\in Child(d)}{f(w;c)},
\end{equation}

\noindent where $f(w;d)$ is the frequency of term $w$ in a document $d$ before propagation, $f^{\prime}(w;d)$ is the frequency of term $w$ in document $d$ after propagation, $c$ is a child page of $d$ in the sitemap $\mathcal{T}$, and $\alpha$ is a parameter to control the mixing factor of the children. This propagation algorithm assumes that the sitemap, $\mathcal{T}$, is constructed ahead of time.

For the purposes of Web information retrieval language models are often used to normalize and smooth word distributions. For illustration purposes, we apply a Dirichlet prior smoothing function~\citep{Zhai2004} to smooth the term distribution where the $f^{\prime}(w;d)$ from above is used in place of the usual $c(w;d)$ from the original Dirichlet prior smoothing function yielding: 

\begin{equation}
    p_{\mu}(w;d) = \frac{f^{\prime}(w;d) + \mu p(w|C)}{|d|^{\prime} + \mu},
\end{equation}

\noindent where $C$ is the distribution over all terms in $V$, $\mu$ is the smoothing parameter, and the length is modified by the propagation algorithm to be $ |d|^{\prime} = (1+\alpha)|d|$. 

As a result of the upward propagation $p_{\mu}$ function, the root document (Web site entry page) will contain all of the words from all of the Web pages in the Web site with different non-zero probabilities. The most probable words are those that occur most frequently and most generally across all documents, and are thus propagated the most. 

\begin{table}
	\centering
	\small{
		\begin{tabular}{c|c|c}
		    $p_{\mu}~\alpha = .5$ & hLDA $\gamma=1$ & HDTM $\gamma=0.95$  \\ \noalign{\hrule height 1.5pt}			 
				
				notre & computer & computer\\ 				
				dame & engineering & engineering \\ 				
				engineering & science & science \\ 				
				university & notre & notre \\ 				
				science & dame & dame \\ 				
				computer & department & department \\ 	\noalign{\hrule height 1.5pt}
		\end{tabular}
		}
	\caption{Comparison of most probable words in top document (in $p_{\mu}$), and in root topic of hLDA and HDTM}
	\label{tab:inh_lm}
\end{table}

As a small, preliminary example, Table~\ref{tab:inh_lm} shows the top six most probable words in the top document (via text propagation) and in root topics of hLDA and HDTM of the computer science department's Web site at the University of Notre Dame\footnote{{\small\url{http://cse.nd.edu}}}. This small example reinforces the intuition that certain Web sites have a hidden hierarchical topical structure. 

In the previous term propagation work, Web site sitemaps were constructed ahead of time using URL heuristics or manually. The goal of HDTM is to learn the document hierarchy automatically and in conjunction with the topical hierarchy.

\subsection{Other hierarchies}
Documents from many different collections exist in hidden hierarchies. While technically a Web site, Wikipedia documents and categories form a unique document graph. Wikipedia categories are especially interesting because they provide a type of ontology wherein categories have more specific sub-categories and more general parent-categories. Most Wikipedia articles are are represented by at least one category description; this allows for users to drill down to relevant articles in a very few number of clicks by browsing the category graph. 

\begin{figure}
	\centering
		\includegraphics[width=\textwidth]{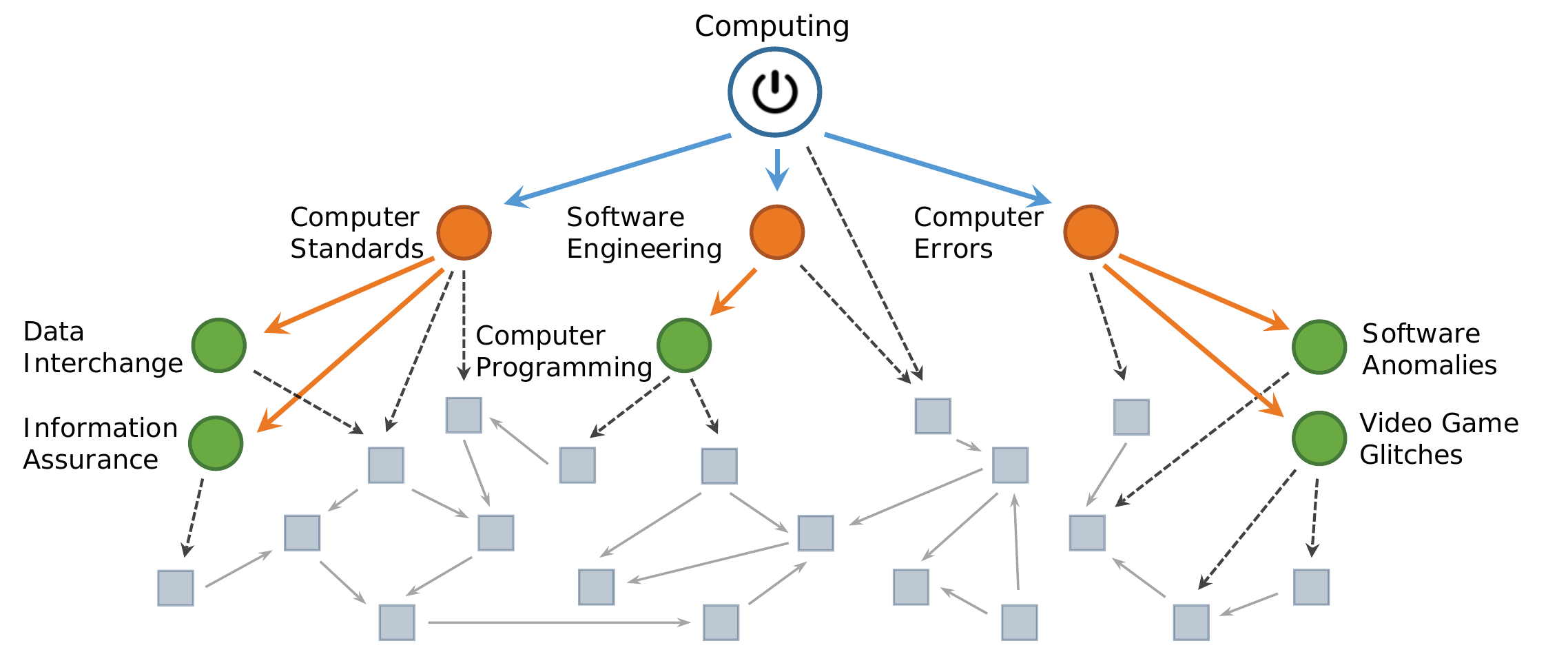}		
	\caption{Truncated portion of the Wikipedia category subgraph rooted at the node \textsf{Computing}. Circles and squares represent Wikipedia categories and articles respectively; Solid, color lines represent category-to-category links, dashed black lines represent category-to-article links, and solid grey lines represent article-to-article links.}
	\label{fig:inh_wikicat}
\end{figure}

A partial example of the Wikipedia category graph is shown in Figure~\ref{fig:inh_wikicat}. This figure illustrates how a document graph can be construed into a document hierarchy. Specifically, the Wikipedia community has hand-crafted a category hierarchy, represented by colored circles and edges, on top of the article-graph, represented by grey squares and grey edges. Although the category graph is imperfect and incomplete, Wikipedia browsers can get a sense of the granularity, topic and fit simply by viewing the article's placement in the category hierarchy. With the Wikipedia construction in mind, our goal can be loosely interpreted as automatically inferring the Wikipedia category hierarchy from the Wikipedia article graph; this inference is tried and evaluated in Section~\ref{sec:inh_experiments}.

Bibliographical networks may also be hierarchically structured. In a bibliographic network, papers or authors (wherein each author could be a collection of documents) are represented by nodes and each citation is represented by an edge in the graph.

Apart from Web and citation graphs, bioinformatics networks, for example protein networks, can also be hierarchically organized for protein fold prediction~\citep{lin2013hierarchical}. The nodes in such network are proteins and two proteins are connected if they have structural or evolutionary relationships~\citep{murzin1995scop}.

\subsection{Challenges and Contributions}

The goal of this work is to construct node hierarchies from an information network using the node features ({\em e.g.}, text) and inter-node edges. For these purposes a node which appears {\em above}, {\em below} or {\em at the same level as} another node refers to the conceptual granularity of the nodes. In other words, given an information network with an explicitly identified root, such as a Web site homepage, we aim to learn a node-hierarchy that best captures the conceptual hierarchy of the document-graph. This problem poses three technical challenges:

\begin{enumerate}
	\item {\bf Document-Topic Inference}. In document hierarchies the parent documents consist of topics that are more general than their children. This requires that the parent documents are viewed as a mixture of the topics contained within its children, and children documents should topically fit underneath their selected parent. In this paper we introduce the {\em Hierarchical Document-Topic Model} (HDTM) which generates a course-to-fine representation of the text information, wherein high-level documents live near the top of the hierarchy, and low-level, more specific documents live at or near the leaves.
	\item {\bf Selecting document placement}. Placement of a document within the hierarchy drives the topic mixing during inference. Because links between edges hint at the context of the relationship between documents, the document placement in the inferred hierarchy is constrained by their edges within the original document-graph. In other words, if an edge exists in the final, inferred hierarchy, then it must also exist in original document-graph (not vice versa). Unlike existing models, such as hLDA~\citep{Blei2004, Blei2010}, that select topic paths using the nested Chinese Restaurant Process (nCRP), HDTM performs document placement based on a stochastic process resembling random walks with restart (RWR) over the original document-graph. The use of a stochastic process over the document-graph frees the algorithm from rigid parameters; perhaps most importantly the adoption of the RWR stochastic process instead of nCRP allows documents to live at non-leaf nodes, and frees the algorithm from the depth parameter of hLDA.
	\item {\bf Analysis at Web site-scale}. In many document-graph collections, the number of edges grows quadratically with the number of nodes. This limits the scalability of many topic diffusion algorithms~\citep{Nallapati2011,Furukawa2008}. An important side-effect of the RWR process is the ability to adapt the HDTM model inference algorithm into a distributed graph processing system that is capable of processing billion-node graphs.
\end{enumerate}

The remainder of this paper is organized as follows: After reviewing the related work we introduce the HDTM model and show how inference is performed. Next we show how the inference can be adapted to a large scale graph processing system in order to run on Web-scale data sets. The experiments section describes a myriad of tests that were run on various data sets and the quantitative and qualitative evaluations that were performed. We conclude with a examples of real-world use cases, discuss avenues for future research, and provide a link to HDTM's source code.

\section{Related Work}

Initial efforts in hierarchical clustering used greedy heuristics such as single-link or complete-link agglomoration rules to infer dendrograms~\citep{Willett1988}, in which the root node is split into a series of branches that terminate with a single document at each leaf. Ho, {\em et al.}, point out that manually-curated Web hierarchies like the Open Directory Project\footnote{\small\url{http://www.dmoz.org}} are typically flatter and contain fewer inner nodes than agglomerative clustering techniques produce~\citep{Ho2012}. Other hierarchical clustering algorithms include top-down processes which iteratively partition the data~\citep{Zhao2005}, incremental methods like CobWeb~\citep{Fisher1987}, Classit~\citep{Gennari1989}, and other algorithms optimized for hierarchical text clustering. 

\begin{figure}
	\centering
\subfigure[hLDA, TopicBlock]{
	\includegraphics[width=.45\textwidth]{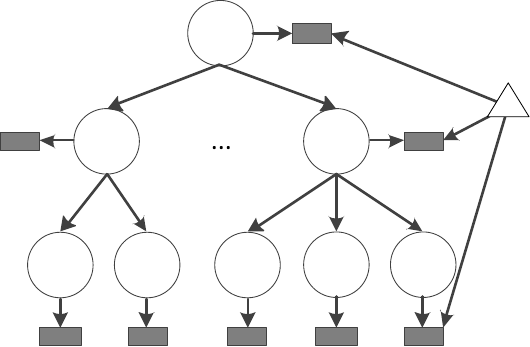}
	\label{fig:rwhlda}
}
\subfigure[hPAM]{
	\includegraphics[width=.45\textwidth]{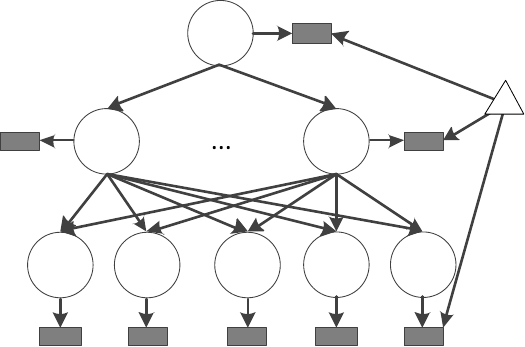}
	\label{fig:rwhpam}
}
\subfigure[TSSB]{
	\includegraphics[width=.45\textwidth]{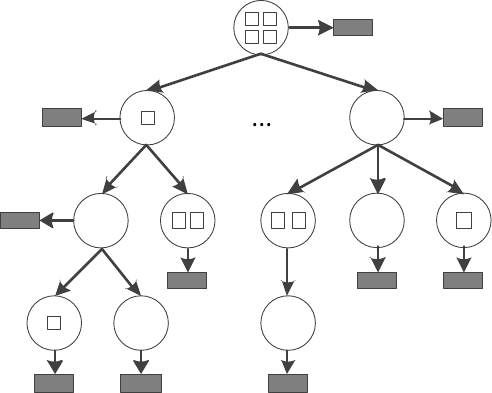}
	\label{fig:rwtssb}
}
\subfigure[\textbf{HDTM}, fsLDA]{
	\includegraphics[width=.45\textwidth]{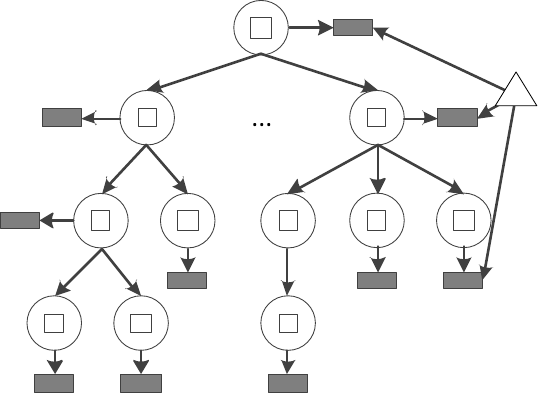}
	\label{fig:rwchlda}
}	
	\caption{Generative structures in related work. Circles and squares represent topics and documents respectively. Each topic has a multinomial over words (grey boxes), and a separate distribution over levels for each path (white triangles). Hierarchical LDA and TopicBlock learn a conceptual taxonomy in which the non-leaf topics are a combination of words and does not represent a real document in the corpus; hPAM constructs a tree-like conceptual taxonomy where each abstract topic has multiple parents; the topics in TSSB are mapped to a document multinomial. HDTM and fsLDA, on the other hand, convert the corpus into document hierarchy by mapping each topic to one, and only one, document. }
	\label{fig:inh_relatedwork}
\end{figure}

The processes that typically defines most hierarchical clustering algorithms can be made to fit in a probabilistic setting that build bottom-up hierarchies based on Bayesian hypothesis testing~\citep{Heller2005}. On the other hand, a lot of recent work uses Bayesian generative models to find the most likely explanation of observed text and links. The first of these hierarchical generative models was hierarchical latent Dirichlet allocation (hLDA). In hLDA each document sits at a leaf in a tree of fixed depth $L$ as illustrated in Figure~\ref{fig:rwhlda}. Note that all non-leave nodes in Figure~\ref{fig:rwhlda} are conceptual topics containing word distribution instead of a document. Each document is represented by a mixture of multinomials along the path through the taxonomy from the document to the root. Documents are placed at their respective leaf nodes stochasically using the nested Chinese restaurant process (nCRP) along side an LDA-style word sampling process. 

NCRP is a recursive version of the standard Chinese Restaurant Process (CRP), which progresses according to the following analogy: An empty Chinese restaurant has an infinite number of tables, and each table has an infinite number of chairs. When the first customer arrives he sits in the first chair at the first table with probability of 1. The second customer can then chose to sit at an occupied table with probability of $\frac{n_i}{\gamma+n-1}$ or sit at a new, unoccupied table with probability of $\frac{\gamma}{\gamma+n-1}$, where $n$ is the current customer, $n_i$ is the number of customers currently sitting at table $i$, and $\gamma$ is a parameter that defines the affinity to sit at a previously occupied table.

The nested version of the CRP extends the original analogy as follows: At each table in the Chinese restaurant are cards with the name of another Chinese restaurant. When a customer sits at a given table, he reads the card, gets up and goes to that restaurant, where he is reseated according to the CRP. Each customer visits $L$ restaurants until he is finally seated and is able to eat. This process creates a tree with a depth of $L$ and a width determined by the $\gamma$ parameter. This process has also been called the Chinese Restaurant Franchise because of this analogy~\citep{Blei2003}.

Adams, {\em et al.} proposed a hierarchical topic model called tree structured stick breaking (TSSB), illustrated in Figure~\ref{fig:rwtssb}, wherein documents can live at internal nodes, rather than exclusively at leaf nodes\citep{Adams2010}. However, this process involves chaining together conjugate priors which makes inference more complicated, and it also does not make use of link data. 

Other work along this line include hierarchical labeled LDA (hL\-LDA) by Petinot {\em et al.}~\citep{Petinot2011} hLLDA, as well as fixed structure LDA (fsLDA) by Reisinger and Pasca~\citep{Reisinger2009} which modify hLDA by fixing the hierarchical structure and learning hierarchical topic distributions. The hierarchical pachinko allocation model(hPAM), shown in Figure~\ref{fig:rwhpam}, produces a directed acyclic graph (DAG) of a fixed depth allowing for each internal (non-document) node to be represented a mixture of more abstract, {\em i.e.}, higher level, topics~\citep{Mimno2007}.

In network-only data, community discovery is the process of finding self-similar group, or clusters. The SHRINK algorithm creates hierarchical clusters by identifying tightly-knit communities and by finding disparate clusters by looking for hubs and other heuristics~\citep{Huang2010}. Clauset, {\em et al}, discover dendrograms by Monte Carlo sampling~\citep{Clauset2008}; however, dendrograms poorly represent the manually curated hierarchies and taxonomies that we are pursuing.

Stochastic block models (SBM) are an alternative line of network clustering research that partitions nodes into communities in order to generatively infer link probabilities~\citep{Holland1983}. Several extensions to the original SBM have since been proposed (for a survey see~\citep{Goldenberg2009}). One downside to block-model processes is that they assign probabilities to every possible edge requiring $\mathcal{O}(N^2)$ complexity in every sampling iteration. Furthermore, SBM methods typically are not concerned with topical/conceptual properties of the nodes.

Because HDTM merges document text and inter-document links into a single model, we assume that the words and their latent topics fit within the link structure of the graph, and that the graph structure explains topical relationships between interlinked documents. Topic Modeling with Network Structure (TMN) is similar in this regard because it regularizes a statistical topic model with a harmonic regularizer based on the graph structure in the data; the result is that topic proportions of linked documents are similar to each other~\citep{Mei2008}. However, hierarchical information is not discovered nor can be easily inferred from this model. 

Topic-sensitive PageRank combines document topics with the PageRank algorithm, by arguing that the PageRank score of a document ought to be influenced by its topical connection to the referring document~\citep{Haveliwala2002}. Like TMN model, Topic-senstive PageRank does not construct any type of information network hierarchy.

Other work on generative models that combine text and links include: a probabilistic model for document connectivity~\citep{Cohn2000}, the Link-PLSA-LDA and Pairwise-Link-LDA methods~\citep{Nallapati2008}, the Latent Topic Model for Hypertext (LTHM) method~\citep{Gruber2008}, role discovery in social networks~\citep{McCallum2005}, the author-topic-model~\citep{Rosen-Zvi2004}, and others. The above models operate by encoding link probability as a discrete random variable or a Bernoulli trial that is parameterized by the topics of the documents. The relational topic model (RTM) builds links between topics, where observed links are given a very high likelihood~\citep{Chang2009b}. The TopicBlock model combines the non-parametric hLDA and stochastic block models~\citep{Holland1983} to generate document taxonomies from text and links~\citep{Ho2012}; however, TopicBlock, like hLDA, does not permit documents to reside at non-leaf nodes of the resulting tree.

To apply topic modeling algorithms on web-scale data, several parallel algorithms have been introduced. Newman, {\em et al} proposed an exact distributed Gibbs sampling algorithm as well as an approximate distributed Gibbs sampling algorithm that uses local Gibbs sampling and global synchronization~\citep{newman2007distributed}. Smyth, {\em et al} introduced an asynchronous distributed algorithm that was capable of learning LDA-style topics~\citep{smyth2009asynchronous}; and Ahmed, {\em et al} recently released Yahoo LDA, which is a scalable approximate inference framework on large-scale streaming data~\citep{ahmed2012scalable}. These parallel approaches use the conventional, procedural programming paradigms, and as a result cannot guarantee statistically sound samples from their Gibbs iterations. Although conventional parallel and distributed algorithms indeed divide the document set into smaller groups, the maximum number of subgroups is subject to the number of processors. In contrast, the vertex-programming paradigm, {\em e.g.}, Pregel, GraphLab and GraphX~\citep{Low:2012wf,Xin:2013bf}, can distribute sampling operations at a much finer granularity by treating each graph-node as an independent computing unit. 

Although the distributed variants of topic inference have made significant contributions in large scale topic models on the one hand, and large scale graph processing on the other hand, we are unaware of any parallel algorithm capable of joining these two subjects to infer topical hierarchies on large scale information networks.

In contrast to the previous work, HDTM builds a hierarchy of documents from text and inter-document links. In this model, each node in the hierarchy contains a single document, and the hierarchy's width and depth is not fixed. The distributed version of proposed algorithm has the ability to handle graph with millions of nodes and billions of tokens.

\section{Hierarchical Document Topic Model}
\label{sec:inh_HDTM}
The problem of inferring the document hierarchy is a learning problem akin to finding the single, best parent for each document-node. Unlike previous algorithms, which discover latent topic taxonomies, the hierarchical document-topic model (HDTM) finds hidden hierarchies by selecting edges in the document graph. This section presents a detailed description of the model. 


Beginning with a document graph $G=\{V,E\}$ of documents $V$ and edges $E$. Each document is a collection of words, where a word $w$ is an item in a vocabulary. The basic assumption of HDTM and similar models is that each document can be generated by probabilistically mixing words from among topics. Distributions over topics are represented by $z$, which is a multinomial variable with an associated set of distributions over words $p(w|z,\beta)$, where $\beta$ is a Dirichlet hyper-parameter. Document-specific mixing proportions are denoted by the vector $\theta$. Parametric-Bayes topic models also include a $K$ parameter that denotes the number of topics, wherein $z$ is one of $K$ possible values and $\theta$ is a $K$-D vector. HDTM, and other non-parametric Bayesian models, do not require a $K$ parameter as input. Instead, in HDTM there exist $|V|$ topics, one for each graph node, and each document is a mixture of the topics on the path between itself and the root document.

In the original LDA model, a single document mixture distribution is $p(w|\theta) = \sum_{i=1}^{K}\theta_i p(w|z=i, \beta_i)$. The process for generating a document is (1) choose a $\theta$ of topic proportions from a distribution $p(\theta|\alpha)$, where $p(\theta|\alpha)$ is a Dirichlet distribution; (2) sample words from the mixture distribution $p(w |\theta)$ for the $\theta$ chosen in step 1.

HLDA is an extension of LDA in which the topics are situated in a taxonomy $T$ of fixed depth $L$. The hierarchy is generated by the nested Chinese restaurant process (nCRP) which represents $\theta$ as an $L$-dimensional vector, defining an $L$-level path through $T$ from root to document. Because of the nCRP process, every document lives at a leaf and the words in each document are a mixture of the topic-words on the path from it to the root.

\subsection{Random Walks with Restart}

The nCRP stochastic process could not be used to infer document hierarchies because the nCRP process forces documents to the leaves in the tree. HDTM replaces nCRP with random walk with restart (RWR) (which is also known as Personalized PageRank (PPR))~\citep{Bahmani2010}. In contrast, random walk with teleportation (aka PageRank) random walks by selecting a random starting point, and, with probability $(1-\gamma)$, the walker randomly walks to a new, connected location or chooses to jump to a random location with probability $\gamma$, where $\gamma$ is called the jumping probability\footnote{Most related works denote the jumping probability as $\alpha$, however, this would be ambiguous with the Dirichlet hyper-parameter $\alpha$.}.

In HDTM, the root node is fixed, either as the entry page of a Web site, by some other heuristic or manually. Therefore, for the purposes of hierarchy inference, the random walker is forced to start and restart at the root node.

\begin{algorithm}[t]
\label{alg:rwr}
\SetKwFunction{G}{G}
\SetKwFunction{T}{T}
\SetKwFunction{Ch}{Ch}
\SetKwFunction{Put}{Put}
\SetKwFunction{RWR}{RWR}
\SetKwInOut{Input}{input}
\SetKwInOut{Globals}{globals}
\SetKwInOut{Output}{output}

\Input{Path Probs. $P$, Current Node $u$, Target $k$, Weight $w$}
\Globals{Graph $\G$, Hierarchy $\T$, Restart Prob. $\gamma$}
\Output{$P$}
\BlankLine
\ForEach(\tcc*[f]{\small{child of $u$ in $\T$}}){$v_i \in \T.\Ch(u)$}{
  \If{$v_i \ne k$}{
    $w \gets w + \operatorname{log}\left(\frac{1-\gamma}{\operatorname{len}\left(\T.\Ch(u)\right)}\right)$\;
    \RWR($P$, $v_i$, $k$, $w$)\tcc*[r]{\small{Recur}}
  }
} 
\If(\tcc*[f]{\small{Edge $u$ to $k$ exists in $\G$}}){$u \rightarrow_{\G} k$}{
  $P.\Put(u, w)$\;
}
\caption{Random Walk with Restart}
\end{algorithm}

Say we wish to find the RWR-probability between some node $u$ and some target node $k$. We model this by a random walker visiting document $u$ at time $t$. In the next time step, the walker chooses a document $v_i$ from among $u$'s outgoing neighbors $\{v|u\rightarrow_{T} v\}$ in the hierarchy $T$ uniformly at random. In other words, at time $t + 1$, the walker lands at node $v_i \in \{v|u\rightarrow_{T} v\}$ with probability $1/deg(u)$, where $deg(u)$ is the outdegree of some document $u\in G$. If at any time, there exists an edge to $k \in \{v|u\rightarrow_{G} v\}$, {\em i.e}, an edge between the current node $u$ and the target node $k$ in the original graph $G$, then we record the probability of that new path possibility for later sampling. Alg.~\ref{alg:rwr} describes this process algorithmically. This procedure allows for new paths from the root $r\leadsto k$ to be probabilistically generated based on the current hierarchy effectively allowing for documents to migrate up, down and through the hierarchy during sampling. 

\subsection{Generating document paths}
\label{sec:generating}
Because a document hierarchy is a tree, each document-node can only have one parent. Selecting a path for a document $d$ in the graph $G$ is akin to selecting a parent $u = Pa(d)$ (and grandparents, etc.) from $\{d|u\rightarrow_G d\}$ in the document graph $G$. HDTM creates and samples from a probability distribution over each documents' parent, where the probability of document $u$ being the parent of $d$ is defined as: 

\begin{equation}
\prod_{t=0}^{\mathrm{dep}_{T}(d)-1}{\frac{1-\gamma}{\mathrm{deg}_{T}(d_t)}},
\end{equation}

\noindent where $d_t$ is the walkers current position at time $t$, $\mathrm{dep}_{T}(d)$ is the depth of $d$ in $T$, and $\mathrm{deg}_{T}(d_t)$ is the outdegree of $d_t$ in the hierarchy $T$. In other words, the probability of landing at $d$ is the product of the emission probabilities from each document in the path through $T$ from $r$ to $d$.

This random walker function assigns higher probabilities to parents that are at a shallower depth than those at deeper positions. This is in line with the intuition that flatter hierarchies are easier for human understanding than deep hierarchies~\citep{Ho2012}. Simply put, the restart probability $\gamma$ controls how much resistance there is to placing a document at successive depths.

Algorithmically, HDTM infers document hierarchies by drawing paths $\mathbf{c}_d$ from the $r$ to the document $d$. Thus, the documents are drawn from the following generative process:

\begin{enumerate}
\item Each document $d \in G$ is assigned a topic $\beta_d \sim $ Dir($\eta$):			
\item For each document $d \in G$:
\begin{enumerate}
  	\item Draw a path $\mathbf{c}_d \sim$ RWR($\gamma$)
  	\item Draw an $L$-dim topic proportion vector $\theta$ from Dir($\alpha$), where $L = $len$(\mathbf{c}_d)$.
  	\item For each word $n \in \{1,\ldots,N\}$:
			\begin{enumerate}
	  		\item Choose topic $z_{d,n}|\theta \sim$ Mult($\theta_d$).
	  		\item Choose word $w_{d,n} | \{z_{d,n}, \mathbf{c}_d, \boldsymbol\beta\} \sim$ Mult($\beta_{\mathbf{c}_d, z_{d,n} }$), where $\beta_{\mathbf{c}_d, z_{d,n}}$ is the topic in the $z$th position in $\mathbf{c}_d$.
			\end{enumerate}	
	\end{enumerate}				
\end{enumerate}

In this generative process hierarchical nodes represent documents {\em and} topics, where internal nodes contain the shared terminology of its descendants.

\begin{figure}
	\centering
		\includegraphics[width=\textwidth]{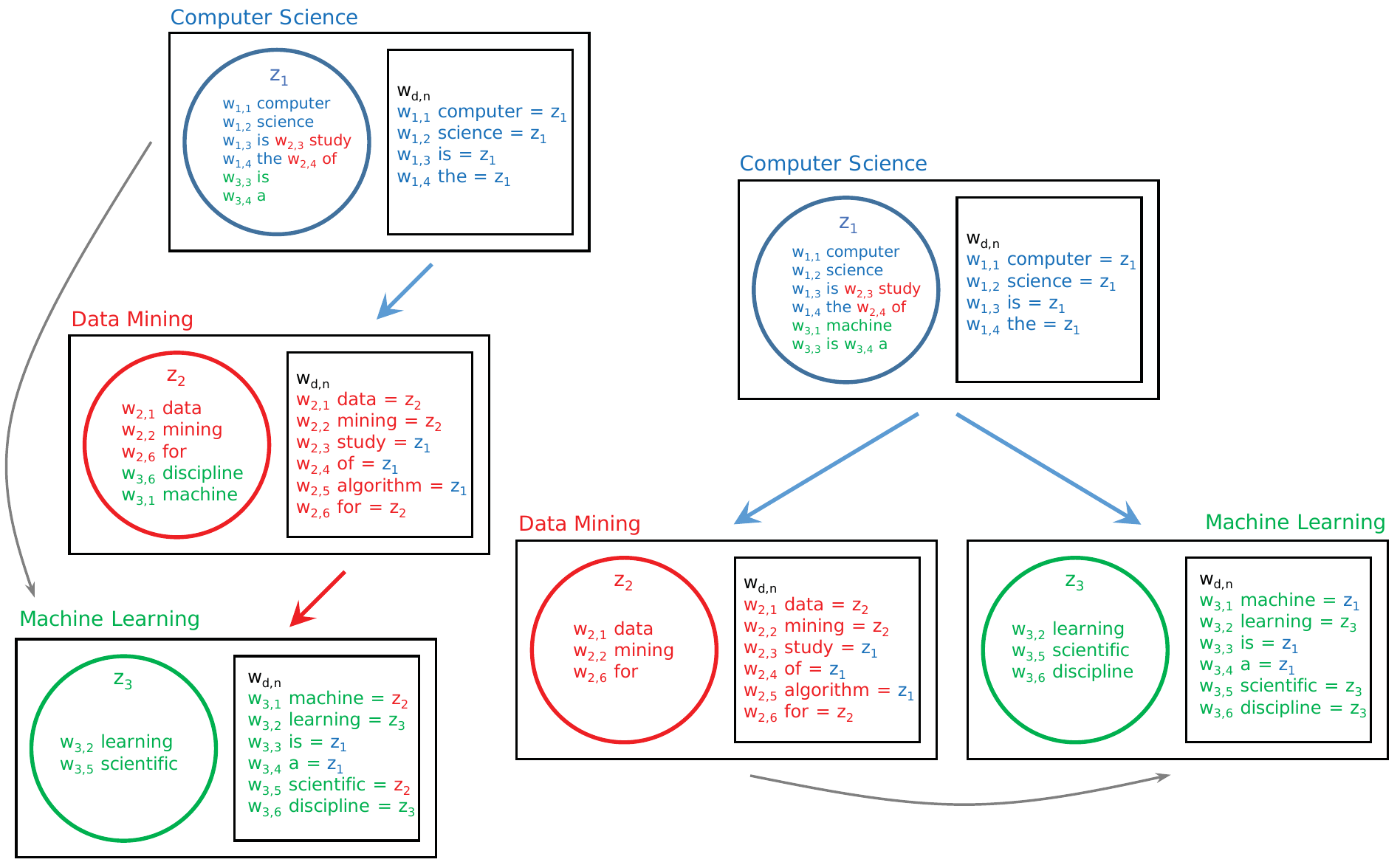}		
	\caption{Illustration of two HDTM samples of the same data. Each node in the hierarchy contains a document and an associated topic. During the generative process, general terms are more likely to be found in topics near the root and vice versa.}
	\label{fig:topicdocuments}
\end{figure}

The illustration in Figure~\ref{fig:topicdocuments} shows the two potential outputs from a three-node graph of the Wikipedia articles: \textsf{Computer Science}, \textsf{Data Mining} and \textsf{Machine Learning}. Clearly, data mining and machine learning could both be regarded as children of computer science, but there is indeed a wiki-link from \textsf{Data Mining} to \textsf{Machine learning} in the graph. There are two key ideas conveyed in this illustration. 

The first is that the hierarchy on the left picks edges so as to ignore the \textsf{Computer Science} to \textsf{Machine Learning} wiki-link (denoted as a thin-grey line), whereas the hierarchy on the right picks edges so as to ignore the \textsf{Data Mining} to \textsf{Machine Learning} wiki-link. Thus, in this toy example, there are two possible hierarchy outcomes presented: the one on the left and the one on the right in Figure~\ref{fig:topicdocuments}.

The second idea conveyed in this illustration is that the word-topic distributions (in colored circles) are stored along with documents' words (in black squares) at the same position in the hierarchy. The word distribution within each topic is constrained such that a document may only propagate terms upwards through the currently selected/sampled hierarchical topology. Thus, it is impossible, in the hierarchy on the left, for the topic distribution at \textsf{Machine Learning} to contain a word unique to \textsf{Computer Science} or \textsf{Data Mining}. It is similarly impossible, in the hierarchy on the right, for the topic distribution in \textsf{Data Mining} to contain a word unique to \textsf{Machine Learning} even though this is indeed possible in the hierarchical topology on the left.

Like in earlier models, there is statistical pressure to have more general terms in topics towards the root of the hierarchy. This is because every path in the hierarchy includes the root node and there are more paths through nodes at higher levels than through nodes at lower levels. Moving down the tree, the topics, and therefore the documents, become more specific.

Hyperparameters also play an important role in the shape and character of the hierarchy. The $\alpha$ parameter affects the smoothing on topic distributions, and the $\eta$ parameter affects the smoothing on word distributions. The $\gamma$ parameter is perhaps the most important parameter because it affects the depth of the hierarchy. Specifically, if $\gamma$ is set to be large ({\em e.g.}, $\gamma = 0.95$) then resulting hierarchy is shallow. Low values ({\em e.g.}, $\gamma = 0.05$) may result in deep hierarchies, because there is a smaller probabilistic penalty for each step that the random walker takes.

\subsection{Inference}
\label{sec:inference}

Exact inference on this model is intractable, so an approximation technique for posterior inference is used. The Gibbs sampling algorithm is ideal in this situation because it simultaneously allows exploration of topic distributions and potential graphical hierarchies. 

The variables needed by the Gibbs sampler are: $w_{d,n}$, the $n$th word in document $d$; $z_{d,n}$, the assignment of the $n$th word in document $d$; and $c_{d,z}$, the topic corresponding to document at the $z$th level. The $\theta$ and $\beta$ variables are integrated out forming a collapsed Gibbs sampler.

The sampling is performed in two parts: (1) given the current level allocations of each word $z_{d,n}$ sample the path $c_{d,z}$, (2) given the current state of the hierarchy, sample $z_{d,n}$. In other words, we use the topic distributions to inform the path selections that make up the hierarchy, and the hierarchy topology to inform the topic distributions.

\subsubsection{Sampling document paths}
The first Gibbs sampling step is to draw a path from each document to the root through the graph. The sampling distribution for a path $c_d$ is
\begin{equation}
\begin{split}
p(c_{d} | \mathbf{c}_{-d}, \mathbf{z}, \mathbf{w}, \eta, \gamma) & \propto p(c_d, \mathbf{w}_d | \mathbf{c}_{-d}, \mathbf{z}, \mathbf{w}_{-d}, \gamma, \eta) \\
& = p(\mathbf{w}_d | \mathbf{c}, \mathbf{z}, \mathbf{w}_{-d}, \eta) p(c_d | \mathbf{c}_{-d}),
\label{eq:pathsample}
\end{split}
\end{equation}
\noindent where $\mathbf{w}$ is the count of terms in document $d$, and $\mathbf{w}_{-d}$ are the words without document $d$. This equation is an expression of Bayes' theorem where the first term represents the probability of data given some choice of path from the root, and the second term represents the probability of selecting some path.

Specifically, the second term represents the probability of drawing the path $c_{d,k}$ to document $d$ at depth $k$ from the RWR process. Recall that each node has an emission probability of $1/\mathrm{deg}_T(d)$, and a restart probability of $\gamma$. The probability is defined recursively:
\begin{equation}
\begin{split}
p(&c_{d,k} | \mathbf{c}_{-d}, c_{d,1:(k-1)})=\prod_{k=0}{\frac{1-\gamma}{\mathrm{deg}_T(d_k)}}
\end{split}
\end{equation}
In other words, the probability of reaching $d$ is equal to the probability of a random walker with restart probability $\gamma$ being at document $d$ at time $k$.

The first term in Eq.~\ref{eq:pathsample} is the probability of a given word based on the current path $\mathbf{c}$ and topic assignment $\mathbf{z}$:
\begin{equation}
\begin{split}
p(\mathbf{w}_{d} | \mathbf{c}, \mathbf{w}_{-d}, \mathbf{z}, \eta ) =&\prod_{k=1}^{\max(\mathbf{z}_d)}{ 
\frac{ \Gamma(\sum_w{ \#[\mathbf{c}_{-d,k} = c_{d,k}, \mathbf{w}_{-d} = w] + W\eta}) }
{ \prod_w{ \Gamma( \#[\mathbf{c}_{-d,k} = c_{d,k}, \mathbf{w}_{-d} = w] + \eta}) } } \times \\
&\quad\hspace{1.1cm} \frac{ \prod_w{\Gamma( \#[\mathbf{z} = k, \mathbf{c}_{k} = c_{d,k}, \mathbf{w} = w] + \eta) } }
{ \Gamma( \sum{ \#[\mathbf{z} = k, \mathbf{c}_{k} = c_{d,k}, \mathbf{w} = w]} + W\eta) }, 
\label{eq:pwgivenc}
\end{split}
\end{equation}

\noindent where $\#[\cdot]$ counts the elements of an array that satisfy the given condition, and $\max(\mathbf{z}_d)$ is the maximum depth of the current hierarchy state. The expression $\#[\mathbf{c}_{-d,k} = c_{d,k}, \mathbf{w}_{-d} = w]$ counts: (ii) $\mathbf{w}_{-d} = w$, {\em i.e.}, the number of words $w$ that do not appear in $d$, for each (i) $\mathbf{c}_{-d,k} = c_{d,k}$, {\em i.e.}, the number of paths to the current document $d$ except those where the path length is $k$. The expression $\#[\mathbf{z} = k, \mathbf{c}_{k} = c_{d,k}, \mathbf{w} = w$ counts: (iii) $\mathbf{w} = w$, {\em i.e.}, the number of words $w$ such that, (ii)  $\mathbf{c}_{k} = c_{d,k}$, {\em i.e.}, the words appear in document $d$ and are situated at the end of a path of length $k$, where (i) $\mathbf{z} = k$, {\em i.e.}, $k$ is one of the topics in $\mathbf{z}$. $W$ is the size of the vocabulary. Eq.~\ref{eq:pwgivenc} is adapted from the standard ratio of normalizing constants for the Dirichlet distribution~\citep{Blei2010}.

\subsubsection{Sampling word levels}
Given the current state of all the variables, the word sampler must first pick an assignment $z$ for word $n$ in document $d$. The sampling distribution of $z_{d,n}$ is
\begin{equation}
\begin{split}
p(z_{d,n} | \mathbf{c}, \mathbf{z}, \mathbf{w}, \eta, \gamma)  & \propto p(w_{d,n}, z_{d,n} | \mathbf{c}, \mathbf{z}_{-(d,n)}, \mathbf{w}_{-(d,n)}, \eta, \gamma) \\ 
&= p(w_{d,n} | \mathbf{c}, \mathbf{z}, \mathbf{w}_{-(d,n)}, \eta)  p(z_{d,n} | \mathbf{z}_{d,-n}, \mathbf{c}, \gamma)
\end{split}
\label{eq:wordsample}
\end{equation}
 where $\mathbf{z}_{d,-n} = \{z_{d,\cdot}\} \setminus z_{d,n}$ and $\mathbf{w}_{-(d,n)} = \{w\} \setminus w_{d,n}$. The first term is a distribution over word assignments:
\begin{equation}
\begin{split}
p(w_{d,n} | \mathbf{c}, \mathbf{z}, \mathbf{w}_{-(d,n)}, \eta) &\propto \#[\mathbf{z}_{-(d,n)} = z_{d,n}, \mathbf{c}_{z_{d,n}} = c_{d, z_{d,n}}, \mathbf{w}_{-(d,n)} = w_{d,n}] + \eta
\end{split}
\end{equation}
which is the $\eta$-smoothed frequency of seeing word $w_{d,n}$ in the topic at level $z_{d,n}$ in the path $c_d$.

The second term is the distribution over levels
\begin{equation}
\begin{split}
p(z_{d,n} = k | \mathbf{z}_{d,-n}, \mathbf{c}, \gamma)
= & \left(\prod_{j=1}^{k-1}{ \frac{1-\gamma}{\mathrm{deg}_T(d_{j-1})} \frac{\#[\mathbf{z}_{d,-n} > j]}{\#[\mathbf{z}_{d,-n} \ge j]} }\right) \times \\
& \frac{1-\gamma}{\mathrm{deg}_T(d_{k-1})} \frac{\#[\mathbf{z}_{d,-n} = k]}{\#[\mathbf{z}_{d,-n} \ge k]},
\end{split}
\label{eq:abuse}
\end{equation}
where $\#[\cdot]$ is the number of elements in the vector which satisfy the given condition. Eq.~\ref{eq:abuse} abuses notation so that the product from $j=1$ to $k-1$ combines terms representing nodes at the $j$th level in the path $\mathbf{c}$ down to the parent of $d_k$, and the second set of terms represents document $d_k$ at level $k$. The $>$ symbol in Eq.~\ref{eq:abuse} refers to terms representing all ancestors of a particular node, and $\ge$ refers to the ancestors of a node including itself.

\subsection{Distributed HDTM}
\label{sec:distinference}

A common complaint among data science practitioners is that graphical models, especially, non-parametric Bayesian graphical models, do not perform well at scale. With this in mind, we also implemented the HDTM inference algorithm in the scalable, distributed vertex-programming paradigm.

The mechanism behind Gibbs sampling, and other Markov Chain Monte Carlo methods, requires sequential sampling steps, and execution of each step depends on the results of the previous step making Gibbs samplers, and MCMC method in general, difficult to parallelize. Approximate Distributed LDA (AD-LDA) is one attempt to find approximate, distributed solutions to the serial inference problem by dividing documents into $P$ parts where $P$ is the number of processors and initializes the topic distribution $z$ globally. Then, for every Gibbs iteration, each processor samples $\frac{1}{P}^\textrm{th}$ of the dataset using the $z_P$ from last Gibbs sampling iteration. When all processors are finished, a global synchronization is performed and $z$ is updated~\citep{newman2007distributed}.

Following the distribution lessons from AD-LDA we sought to also implement a scalable, distributed version of HDTM. However, a major difference between the LDA/hLDA and HDTM is that hLDA uses the nCRP stochastic process to assign terms to topics, while HDTM samples paths from a graph of documents using graph-based random walk with restart method. The process of random walks over the network topology combined with the term sampling process described above is a good candidate for the vertex-programming paradigm using frameworks like Pregel ~\citep{Malewicz2010} or GraphLab~\citep{Low:2012wf}.

\subsubsection{Vertex Programming}

Although MapReduce is a widely used, general purpose parallel scheme that can easily deal with scalable data, it is not optimized for \emph{iterative} computational tasks such as statistical inference or logistic regression~\citep{zaharia2010spark}. This is because MapReduce materializes all intermediate results and to disk in order to tolerate task failures. Mapreduce, therefore, has relatively high I/O costs compared to other designs that keep data in memory across iterations~\citep{Lee:2012:PDP:2094114.2094118}. 

Apart from MapReduce, another scalable solution is to build a custom distributed system using message passing interface (MPI). Custom approaches are usually closer to optimal because developers can tune the code based on their own needs and minimize the unnecessary overhead. However, the drawbacks are also significant: because MPI is a barebone communication specification developers need to write their own code for job dispatching, load balancing, and dealing with node failure~\citep{zou2013survey}. 

After evaluating aforementioned approaches, we decided to use an emerging computational paradigm called vertex programming. Vertex programming aims to improve the performance of graph/network computing by automatically distributing in-memory computation and with vertex-centric scheduling. Unlike the MapReduce paradigm, which writes every intermediate result to disk, vertex programming keeps data up-to-date, in-memory and reduces I/O overhead by only materializing data to disk through periodic checkpoints. Vertex-centric scheduling views every graph-vertex as an elementary computing unit and uses MPI (or some other message passing system) to transfer data over graph-edges. Typically, vertex-programs are rather easy to implement, can be distributed easily, and are much more computationally efficient than conventional, procedural programming when working with iterative computational tasks~\citep{McCune2015}.

Several vertex programming frameworks, including Spark-GraphX, HAMA, GraphLab, GraphLab Create and GraphLab 2.0 (PowerGraph), were evaluated, and the PowerGraph framework was ultimately chosen. The decision was based, in part, from insurmountable troubles experienced during several months of implementation attempts on Spark's GraphX framework. Ultimately, we concluded that C++ and MPI based PowerGraph was faster, more scalable and had a much smaller memory footprint than other frameworks. 

\subsubsection{Distributed Inference Algorithm}

\begin{algorithm}[t]
\label{alg:dist_rwr}
\SetKwFunction{G}{G}
\SetKwFunction{T}{T}
\SetKwFunction{Ch}{Ch}
\SetKwFunction{Sendmsg}{sendmsg}
\SetKwFunction{Getmsg}{getmsg}
\SetKwFunction{sum}{sum}
\SetKwInOut{Input}{input}
\SetKwInOut{Globals}{globals}
\SetKwInOut{Output}{output}

\Globals{Vertices $V$, Hierarchy $T$, Restart Prob. $\gamma$}
\Input{Messages received containing path probabilities $m$}
\Output{Messages sent to adjacent edges}
\BlankLine

\ForPar(\tcc*[f]{each document in parallel}){vertex $d_k \in V$}{
    \If{$\textbf{m}\gets \Getmsg() $ }{
        $c_{d,k} \gets c_{d,k} + \sum(m)$ \tcc*[f]{\small{Add path-probs from incoming message $m$}}\\
        \ForEach{$child \in \T.\Ch(d_k)$}{
                \tcc*[f]{\small{Send prob message to children of $d_k$}}\\
            \Sendmsg($child$, $c_{d,k} + \operatorname{log}\left(\frac{1-\gamma}{\operatorname{len}\left(\T.\Ch(d_k)\right)}\right)$) \\
        }
    }
}
\caption{Distributed Random Walk with Restart}
\end{algorithm}

\begin{algorithm}[]
\label{alg:dist_upate}
\SetKwFunction{G}{G}
\SetKwFunction{T}{T}
\SetKwFunction{Path}{Path}
\SetKwFunction{Update}{update}
\SetKwFunction{Getmsg}{getmsg}
\SetKwFunction{Sendmsg}{sendmsg}
\SetKwInOut{Input}{input}
\SetKwInOut{Globals}{globals}
\SetKwInOut{Output}{output}

\Globals{Vertices $V$, Hierarchy $T$, Restart Prob. $\gamma$}
\Input{Messages received containing path probabilities $m$}
\Output{Messages sent to adjacent edges}
\BlankLine

\ForPar(\tcc*[f]{\small{each document in parallel}}){vertex $d_k \in V$}{
    \ForEach{$u \in c_{d,k}$}{
        \Sendmsg($u, [d_k.n, d_k.z]$) \tcc*[f]{\small{Send local $n$, $z$ to node $u$}}\\
    }
    \If{$\textbf{m} \gets \Getmsg()$}{
        \ForEach{$u \in \textbf{m}$}{
            $d_k.n \gets u.n$\\
            $d_k.z \gets u.z$\\
        }
    }
}
\caption{Path-Global Update}
\end{algorithm}

The distributed HDTM inference algorithm is similar to procedural HDTM. We do not detail the entire distributed HDTM inference algorithm in this paper; however, the source code is referenced in the Section~\ref{sec:conclusions}. To fit HDTM to the vertex-programming model, changes in sampling sequence and attention to global synchronization were required. Firstly, random walk with restart must be executed during each Gibbs iteration so that every visited node can have a random walk probability. Next, every node gathers $\mathbf{w}_{-d}$, $\mathbf{z}$, and $\mathbf{c}$ separately and decides their new path $\mathbf{c}$ according to Eq~\ref{eq:pathsample} and shown in Alg~\ref{alg:dist_rwr}. After a path is sampled, each node will pick sample assignments $z$ for each word $n$ across all documents/nodes $d$ in parallel according to Eq~\ref{eq:wordsample}. A global synchronization step is required so that each document/node can update the number of words $n$ and the topic assignments $z$ globally; fortunately, because the topics and words are sampled according to the information provided in the {\em path} from root $r$ to the each document $d_k$ (document $d$ at level $k$), it suffices to update the nodes on the path from $r$ to $d_k$ instead of an actual global update to all nodes in $V$. This vertex-programming based path-global update function is shown in Alg~\ref{alg:dist_upate}. Furthermore, this update is executed during the {\em synchronization barrier}, which is built-in to most vertex-programming frameworks, is highly optimized, and does not lock the global system any more than the synchronization barrier already does. 

\section{Experimental Results}
\label{sec:inh_experiments}
This section describes the method and results for evaluating the HDTM model. A quantitative and qualitative analysis of the hierarchical document-topic model's ability to learn accurate and interpretable hierarchies of document graphs is shown. The main evaluations explore the empirical likelihood of the data and a very large case study wherein human judges are asked to evaluate the constructed hierarchies.

\begin{table}
	\centering
	\small{
		\begin{tabular}{l|c c c c}
		    		    	& Wikipedia (cat) & Wikipedia (article) & CompSci Web site  & Bib. Network \\ \noalign{\hrule height 1.5pt}	
				documents 	& 609 		&  1,957,268 	& 1,078 						     & 4,713 \\ 				
				tokens		& 5,570,868 &  1,316,879,537     & 771,309 					         & 43,345\\ 				
				links		& 2,014 	&  44,673,134	& 63,052 						     & 8,485\\ 				
				vocabulary 	& 146,624 	&  4,225,765     & 15,101 						     & 3,908\\ \noalign{\hrule height 1.5pt}
		\end{tabular}
		}
	\caption{Comparison of most probable words in top document (in $p_{\mu}$) and in root topic (in hLDA)}
	\label{tab:topdoc}
\end{table}

\subsection{Data}
HDTM is evaluated on four corpora: the Wikipedia category graph, the Wikipedia document graph, a computer science Web site from the University of Illinois, and a bibliographic network of all CIKM and SIGIR conferences. 


The Wikipedia data set has been used several times in the past for topic modeling purpose. However, because of the computational resources needed to infer topic models, prior studies have severely constricted the dataset size. Gruber {\em et al.}, crawled 105 pages starting with the article on the NIPS conference finding 799 links~\citep{Gruber2008}. Ho {\em et al.} performed a larger evaluation of their TopicBlock model using 14,675 document with 152,674 links; however, they truncated each article to only the first 100 terms and limited the vocabulary to the 10,000 most popular words~\citep{Ho2012}. 

The Wikipedia category data set is a crawl of the {\em category} graph of Wikipedia, beginning at the category \textsf{Computing} as shown in Figure~\ref{fig:inh_wikicat}. In Wikipedia each category has a collection of articles and a set of links to other categories; however, categories don't typically have text associated with them, so the text of each article associated with a particular category is associated to the category's text. For example, the category \textsf{Internet} includes articles: \textsf{Internet, Hyperlink, World Wide Web}, etc. In total the crawled category graph consisted of 609 categories with text from 6,745 articles. The category graph is rather sparse with only 2,014 edges between categories, but has vocabulary size of 146,624 with 5,570,868 total tokens. We did not perform any text preprocessing procedure, including stop word removal and stemming, in any of the experiments due to empirical findings that these models are robust to the presents of stop words, etc~\citep{Mccallum2009}. Such settings can also explore HDTM's robustness in presence of a large and noisy corpus.

A computer science department Web site from the University of Illinois Urbana-Champaign was chosen as the second data set because it represents a rooted Web graph with familiar topics. By inferring the document hierarchy, the goal is to find the organizational structure of the computer science department. The intuition is that Web sites reflect the business organization of the underlying entity; thus it is expected that subtrees consisting of courses, faculty, news, research areas, etc. are found at high levels, and specific Web pages are found at lower levels in the hierarchy. Web site was crawled starting at the entry page and captured 1,078 Web pages and 63,052 hyperlinks. In total there were 15,101 unique terms from 771,309 tokens.

The bibliographic network consists of documents and titles from 4,713 articles from the SIGIR and CIKM conferences. There exist 3,908 terms across 43,345 tokens in the document collection. In this collection, links include citations between papers within the CIKM and SIGIR conferences. Due to different citation styles, vendor abbreviations, and ambiguous names, without human interference most bibliographical meta-data extraction algorithms can only extract a portion of correct data~\citep{giles1998citeseer,ley2002dblp,Tang2008}. Here we choose to use the most complete citation data set available from by the authors of the ArnetMiner project~\citep{Tang2008}, however the citation graph is not guaranteed to be complete. A SIGIR 1998 paper by Ponte and Croft~\citep{Ponte1998} was chosen to be the root document because, in the data set, it had the most in-collection citations.

To construct a larger data set with finer granularity, the English Wikipedia article graph is used. In Wikipedia, every Wiki-article has two different link types: 1) category links, which point from the article to its categories, and 2) Wiki-links, which point from the article to other Wiki-articles. The category links are important because they hint at some encoded hierarchical structure that was codified on top of Wikipedia after the article-graph was largely developed. For example, the Wiki-article \textsf{Barack Obama} has a category link to \textsf{Harvard Law School alumni} and Wiki-links to \textsf{Harvard Law School} and \textsf{Dreams from My Father} among others. In later experiments we infer a document hierarchy from Wikipedia and compare it to the crowd-encoded Wikipedia category hierarchy. Our Wikipedia snapshot has 4,606,884 documents with 1,878,158,318 tokens, 44,739,242 links, and a vocabulary of 6,064,216 words. Because the document graph is not a single connected component, we picked the largest component in that graph as data set. This giant component contains 1,957,268 documents and 44,673,134 document to document links; 42.49\% of nodes and 70\% of edges are in this component. Additionally,  1,316,879,537 terms with 4,225,765 unique tokens appear in this component, which is approximately 70\% of the total number of terms. As a graph-preprocessing step, we replaced all edges pointing to redirection pages with edges that point directly to the actual Wiki-article.

\subsection{Summary of the Sampling Algorithm}
Using either the traditional inference model from Sec.~\ref{sec:inference} or the high-throughput distributed sampling algorithm from Sec.~\ref{sec:distinference} we basically sample a full hierarchy $\textbf{c}_d$ by sampling paths for each node $c_d$ except the root and the words in the document $z_{d,n}$. Given the state of the sampler at some time $t$, \ie, $\textbf{c}_{1:D}^{(t)}$ and $\textbf{z}_{1:D}^{(t)}$, we iteratively sample each variable conditioned on the others as illustrated in~\ref{sec:generating}. 

The conditional distribution of the latent variables in HDTM model the given document-network. After running the Markov chain a sufficient number of times we approach the stationary distribution; the process of approaching the stationary distribution is called the ``burn-in'' period. After the burn-in we collect samples at a selected interval, \ie, the sampling lag. 

The collected ``samples'' from the Markov chain are full hierarchies constructed from the selection of a path for each node $c_d$ and a word for each document $z_{d,n}$. Therefore each sampled hierarchy contains one estimation about position of each document in the hierarchy and the position of each word in a document. For a given sampled hierarchy, we can assess the goodness of the hierarchy by measuring the log probability of that hierarchy and the observed words conditioned on the hyperparameters:

$$
\mathcal{L}^{(t)} = \log p\left(\textbf{c}_{1:D}^{(t)}, \textbf{z}_{1:D}^{(t)}, \textbf{w}_{1:D} | \gamma, \eta \right).
$$

Using this log likelihood function it is possible to pick the most sampled hierarchy that maximizes the log likelihood as the final result. Later, we will relax this assumption and use a more robust measurement that computes the mode of the sampled hierarchies and determines the \emph{certainty} of the result.

\subsection{Quantitative Analysis}

\begin{figure}
\centering
\subfigure[The complete log likelihood]{
\includegraphics[width=.45\textwidth]{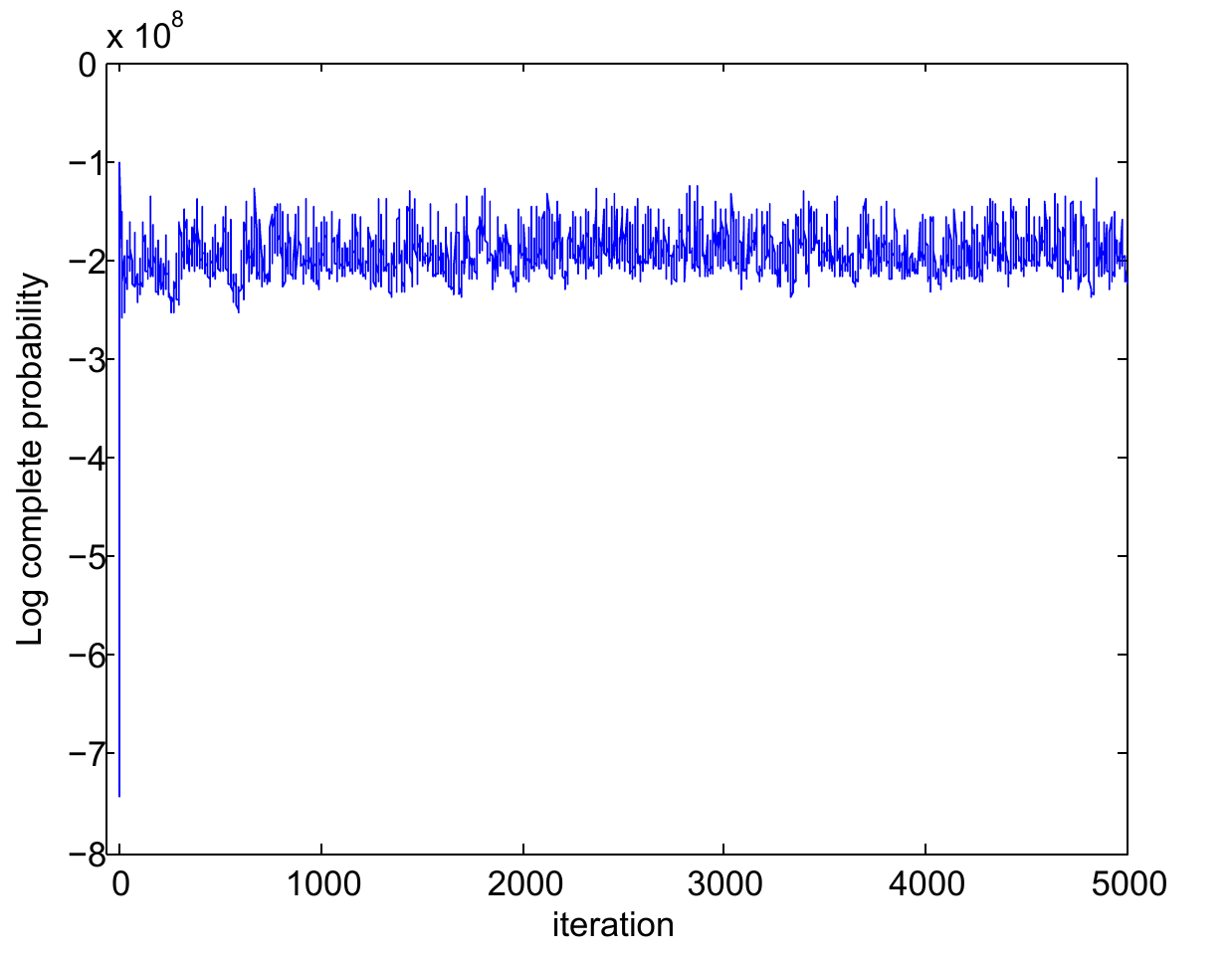}
\label{fig:lliterations}
}
\hspace{.05cm}
\subfigure[Log likelihood and corresponding average node depth. High likelihood corresponds with hierarchies of high average depth.]{
\includegraphics[width=.45\textwidth]{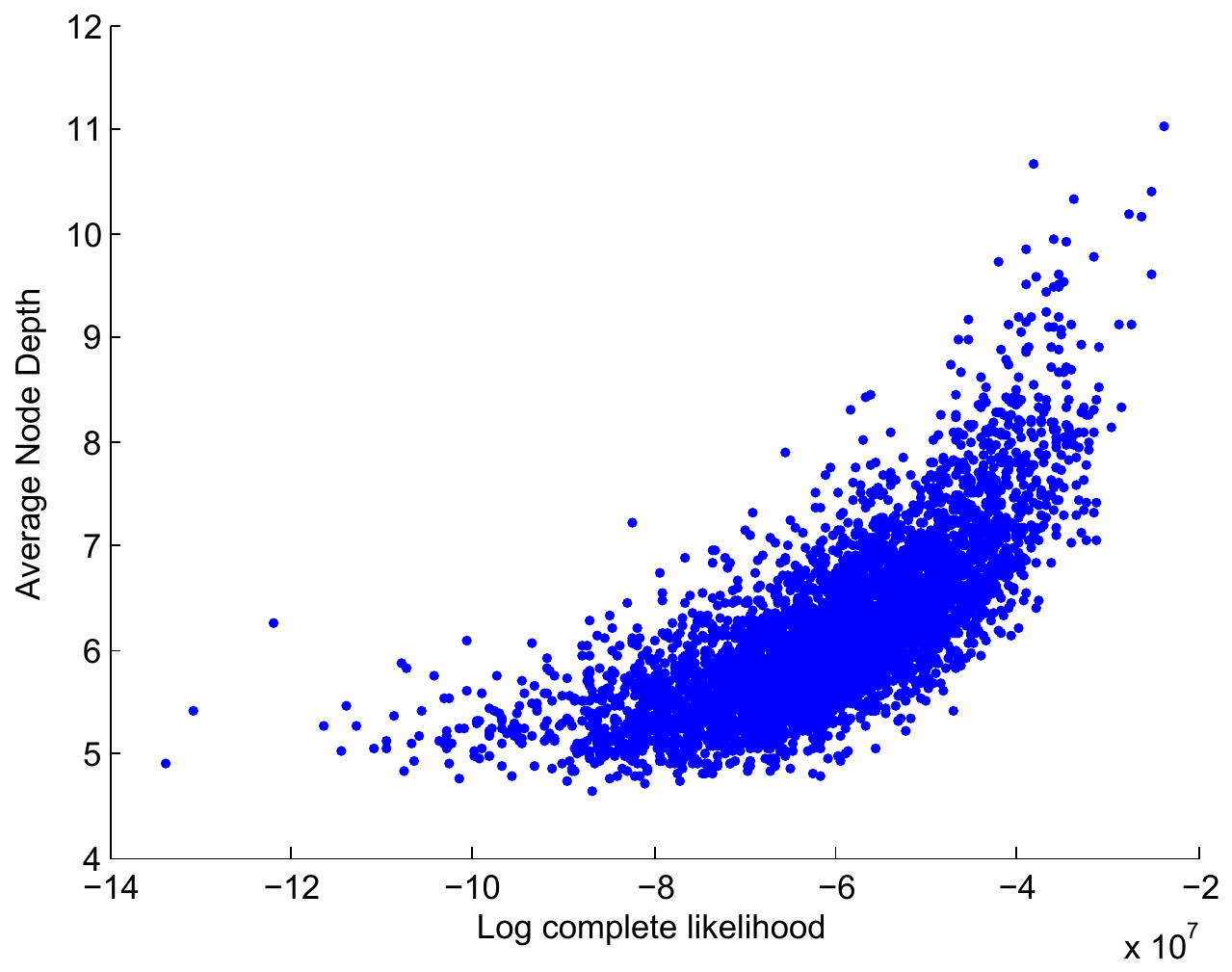}
\label{fig:llavgdepth}
}
\caption{Analysis of log likelihood Scores for 5,000 iterations of HDTM's Gibbs sampler run on the Web Site collection.}
\label{fig:quantanalysis}
\end{figure}

HDTM has some distinct qualities that make apples to apples comparison difficult. Because HDTM is the first model to generate document hierarchies based on graphs, there is nothing to {\em directly} compare against. However, some of the models in the related work perform similar tasks, and so comparisons were performed when applicable.

The related models typically perform quantitative evaluation by measuring the log likelihood on held out data or by performing some other task like link prediction. Log likelihood analysis looks at the goodness of fit on held out data. Unfortunately, the creation of a ``hold out'' data set is not possible, because each document, especially documents on the first or second level document, is very important to the resulting hierarchy. Removing certain documents might even cause the graph to separate, which would make hierarchy inference impossible. Instead, for the first quantitative evaluation, we compare the highest log likelihood generated from each model's sampler.

Quantitative experiments were performed on many of the aforementioned algorithms including: HLDA~\citep{Blei2004, Blei2010}, TopicBlock \citep{Ho2012}, TSSB~\citep{Adams2010}, and fsLDA~\citep{Reisinger2009}. The fixed structure in fsLDA is determined by a breadth first iteration over the document graph because URL heuristics used in the original paper do not exist in most of our data sets. The depth of HLDA and TopicBlock is 4.

In all cases, a Gibbs sampler was run for 5,000 iterations; with a burn-in of 2,000 and a sampling lag of 20. Figure~\ref{fig:lliterations} shows the log likelihood for each sample on the computer science Web site; the other data sets showed similarly-shaped results. Interestingly, Figure~\ref{fig:llavgdepth} shows that higher likelihood values are strongly correlated with hierarchies of deeper average depth.

\begin{figure}
    \centering
    \includegraphics[width=.65\textwidth]{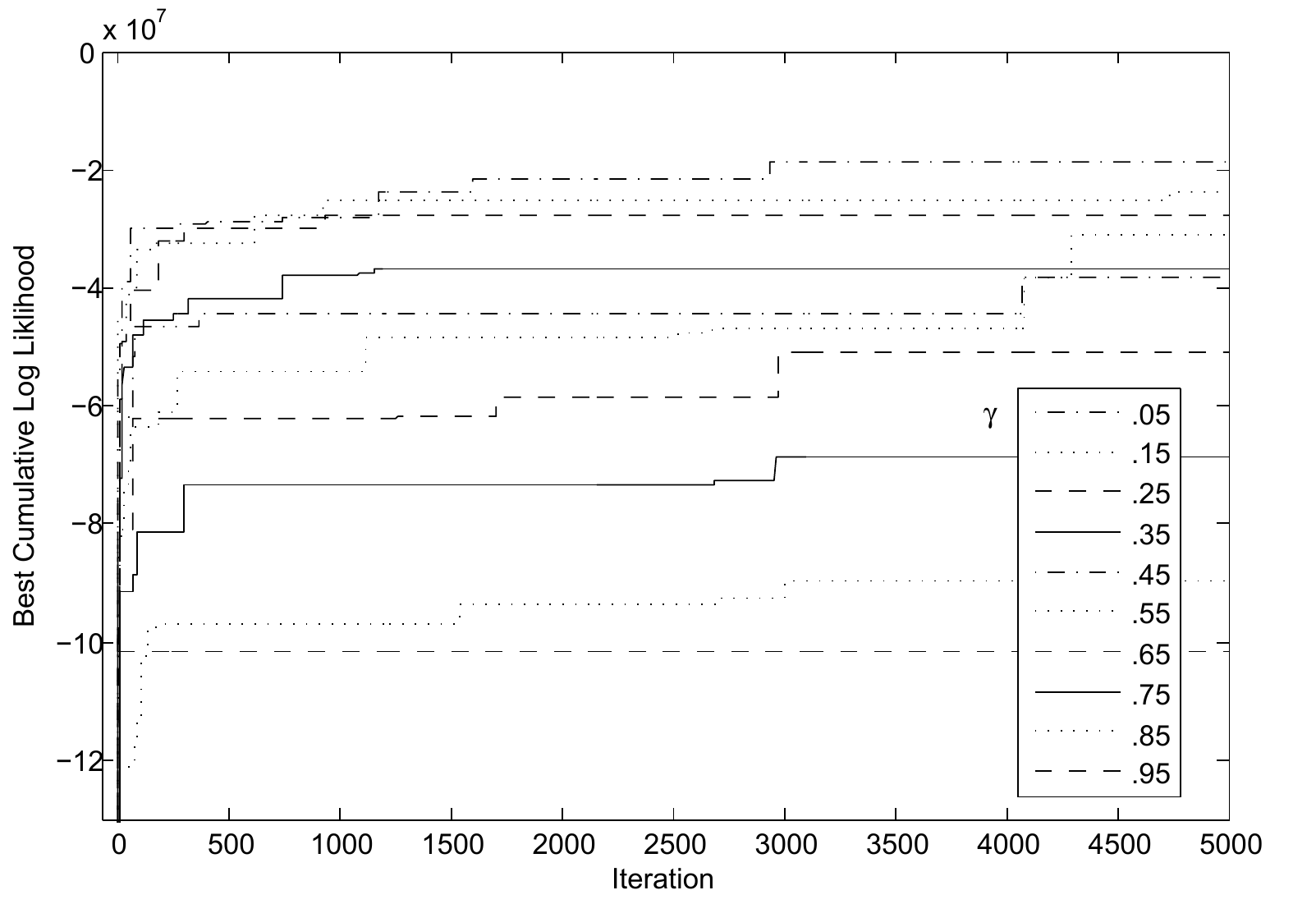}
    \caption{Best cumulative log complete likelihood for each tested $\gamma$ value. Lower $\gamma$ values result in deeper hierarchies.}
    \label{fig:bestcumLL}
\end{figure}

The Gibbs sampling algorithm was run on HDTM for various values of $\gamma$, and Figure~\ref{fig:bestcumLL} shows the best cumulative log likelihood for each of the tested values of $\gamma$. We observe that HDTM with $\gamma = 0.05$ achieved the best likelihood score. Likelihood scores decreased steadily for increasing $\gamma$ values, and HDTM with $\gamma = 0.95$ achieved the worst likelihood score. 

\begin{table}
	\centering
	\small{
		\begin{tabular}{c|c c c}
		      												& CompSci Web site 	& Wikipedia (Cat)		& Bib. Network \\ \noalign{\hrule height 1.5pt}			 				
			HDTM $\gamma=0.05$						& -1.8570					& -148.071		& -0.4758 \\
			HDTM $\gamma=0.95$			 			& -9.2412					& -148.166	  & -0.5183 \\
			HLDA $\gamma=1.0$							& -8.5306					& -50.6732	  & -8.5448 \\
			TopicBlock $\gamma=1.0$				& \textbf{-0.2404}					& -2.9827	    & \textbf{-0.4192} \\
			TSSB $k=10$										& -0.5689					& \textbf{-0.0336}	    & -0.4655 \\
			fsLDA													& -48.9149				& -149.622	  & -0.6602 \\ \noalign{\hrule height 1.5pt}
		\end{tabular}
		}
	\caption{Log likelihood results of the best sample from among 5,000 Gibbs iterations. Values are $\times 10^6$. Higher values are better. Best results are in bold.}
	\label{tab:LLResults}
\end{table}

Table~\ref{tab:LLResults} shows the results of the different algorithms on the three smaller data sets. The TopicBlock and TSSB clearly infer models with the best likelihood. The remaining algorithms, including HDTM, have mixed results. Although these log likelihood results initially seem as if HDTM performs poorly, we demonstrate in Section~\ref{subsec:discussion} that these likelihood results are expected and preferred.

\subsubsection{Large-scale Analysis}

Recall that the goal of HDTM is to generate document hierarchies from some original document-graph. Thus, a positive result should somehow demonstrate that HDTM demonstrates hierarchies that are similar to the hierarchy that a human (or many humans) would create given the same document graph. With this in mind we performed a large scale experiment on the full English Wikipedia article graph. Also recall that the Wikipedia category graph establishes a hierarchy that was human-built on top of Wikipedia that maintains a pseudo-hierarchical classification of almost every document. The structure of categories is described to be a {\em tree} rooted by \textsf{Main topic classifications} or \textsf{Fundamental categories}; however, we find that 87.01\% of categories have an in-degree larger than 1 indicating the actual category is more complex. We find that although the Wikipedia category graph {\em mostly} resembles a top-down (coarse to fine) hierarchy, it is technically an undirected loopy graph. As for Wiki-articles, they almost always belong to more than one category because the idea or object described in each article can rightfully have multiple classification perspectives. For example, the Wiki-article \textsf{Computer Science} belongs to a category of the same name, but it also belongs to \textsf{Computer Engineering} and \textsf{Electrical Engineering} among others. 

For a large scale evaluation our task is to infer a hierarchy from the full Wikipedia article graph and see how well it matches the category graph. Unfortunately, it is difficult to measure similarity between category graph and HDTM-hierarchies using conventional evaluation methods like the log likelihood. Consider, for example, the category \textsf{University of Notre Dame faculty}, which contains Wiki-articles of faculty members from vastly different topical fields. In such cases, the bond is described primarily by the links rather than the content; nevertheless, the log likelihood score judges models based on the goodness of fit of words. Thus it is unreasonable to measure hierarchy quality by likelihood scores alone; indeed, the hierarchy with the best possible likelihood score may be very different from actual, human created hierarchy. In order to perform a more thorough evaluation we should also compare the the graph structure.

Graph topologies can be compared in many different ways. Among many options we chose DeltaCon~\citep{Faloutsos:2013dn} to measure the similarity between HDTM's hierarchy and the Wikipedia category hierarchy three reasons: 1) DeltaCon can calculate the similarity between two graphs with nodes that partially overlap, as observed in the messy Wikipedia category graph, 2) the underlying metric in DeltaCon is the affinity score calculated by fast belief propagation (FABP), which models the connectivity of nodes, and 3) DeltaCon can process millions of nodes in only a few hours. The DeltaCon metric returns a similarity score between the two compared graphs ranging between 0 and 1 inclusive where 0 is completely dissimilar and 1 is completely similar. 

Although comparing with hLDA and its variants is important, all existing hierarchical topic models are incapable of processing on data sets at this scale. We attempted to use the hLDA implementation in the Mallet package~\citep{Mccallum:2002fe}, but 130 hours was needed in order to compute a single Gibbs sampling iteration. Fortunately, we were able to use Graphlab's implementation of LDA which is based on models developed by Ahmed {\em et al.}~\citep{ahmed2012scalable} and Smyth {\em et al.}~\citep{smyth2009asynchronous}. Although LDA does not directly generate a graph, we are able to compare it with the category hierarchy by connecting the documents within a single LDA-cluster to a made-up root node to simulate a hierarchy. The large scale models were inferred using 400 Gibbs sampling iterations with a burn-in of 100 and a lag of 10.

In addition to HDTM and LDA, we also use DeltaCon to compare the original Wikipedia article graph and a randomly generated article-hierarchy as baselines. In the case of Wikipedia article graph, we expect the Wikipedia article graph (the full graph, not a hierarchy) to perform reasonably well because the category hierarchy is largely built on top of the article graph topology and they currently coexist and coevolve. In the case of the randomly generated article hierarchy, we randomly select a single parent for each node in the original article graph.

\begin{table}
\centering
\begin{tabular}{r | c}
		     & DeltaCon \\ \noalign{\hrule height 1.5pt}			 				
 HDTM ($r$=\textsf{Science}, $\gamma=0.05$) & 0.046852 \\
 HDTM ($r$=\textsf{Science}, $\gamma=0.95$) & 0.046851 \\
 HDTM ($r$=\textsf{Barack Obama}, $\gamma=0.05$) & 0.046857 \\
 HDTM ($r$=\textsf{Barack Obama}, $\gamma=0.95$) & 0.046856 \\
 Random Article Hierarchy & 0.046700 \\
 Article Graph & 0.044208 \\
 LDA ($k=10$) & 0.026770 \\
 LDA ($k=50$) & 0.037949 \\
\end{tabular}
\caption{Comparison of the Wikipedia category graph to other generated hierarchies. DeltaCon scoring means higher is better. The random article hierarchy is generated by randomly picking one parent among all possible parents, \ie, references, for each article. The differences in DeltaCon scores show that HDTM can preserve crucial connectivity information when constructing a new hierarchical structure from the original graph.}
\label{tab:deltacon}
\end{table}

Table~\ref{tab:deltacon} shows that the hierarchy inferred by HDTM is indeed most similar to the Wikipedia category graph, followed by the random article hierarchy. Recall that DeltaCon looks at graph topology similarity, thus the HDTM hierarchy is expected to more topologically similar to another hierarchy (even if random) than the article graph. These results demonstrate that HDTM can identify and preserve critical topological features when inferring a hierarchy from the graph. As usual, an increase in the number of LDA topics increases the goodness of fit score (in this case measured by DeltaCon instead of likelihood). Interestingly, the $\gamma$ parameter and the selection of different roots did not significantly influence DeltaCon score. To understand why this is, recall that different root-nodes and $\gamma$ values result in different classification perspectives, but the different perspectives are still subsets of the same category graph with similar topological properties. Hence they should have similar scores using the DeltaCon metric.

 
Recall that the HDTM, in its most basic form, generates a hierarchy by picking the best parent for each node/document (except the root). In the iterative Gibbs sampling process it is an almost certainty that a given node will pick different parents during different iterations. For example, say node $d$ has two parents $x$ and $y$, it is possible that during iterations 1--5 node $d$ samples node $x$ as its parent, but then in iterations 6--20 node $d$ samples node $y$ to be its parent. Two questions come to mind: 1) which parent should be ultimately picked for node $d$ in the final output graph? and 2) what can the distribution of samples say about the certainty of our inferred graph?

In classic Gibbs sampling, maximum a posteriori estimation is determined by the the mode of the samples. For some node $d$, this translates to choosing the parent that appeared in the most samples as the final parent; in the above case node $d$ would chose node $y$ as its parent because node $y$ was sampled 15 times compared to only 5 samples of node $x$. To answer the second question, and therefore model the certainty of an inferred hierarchy, we calculate the number of times some node $d$ picks its final parent over the total number of samples normalized by the number of possible choices, {\em i.e.}, parents, node $d$ has. This results in a certainty score for node $d$:

\begin{equation}
 certainty_d = \frac{\frac{n_{p}}{n} - \frac{1}{deg^{-}(d)}}{\frac{n_{p}}{n}} , 
\end{equation}

where $n$ represents the total number of samples, $n_{p}$ is the number of times the final parent $p$ is sampled, and $deg^{-}(d)$ is the indegree of node $d$ representing the total number of parents that node $d$ could pick from. In the certainty score the outside fraction is used to measure the normalized difference between the raw probability and probability of random guess. Applying this function to the above example we would calculate that the certainty score of node $d$ would be $(\frac{15}{20}-\frac{1}{2}) / \frac{15}{20} = .33$.

\begin{figure}
	\centering
	\includegraphics[width=.40\textwidth]{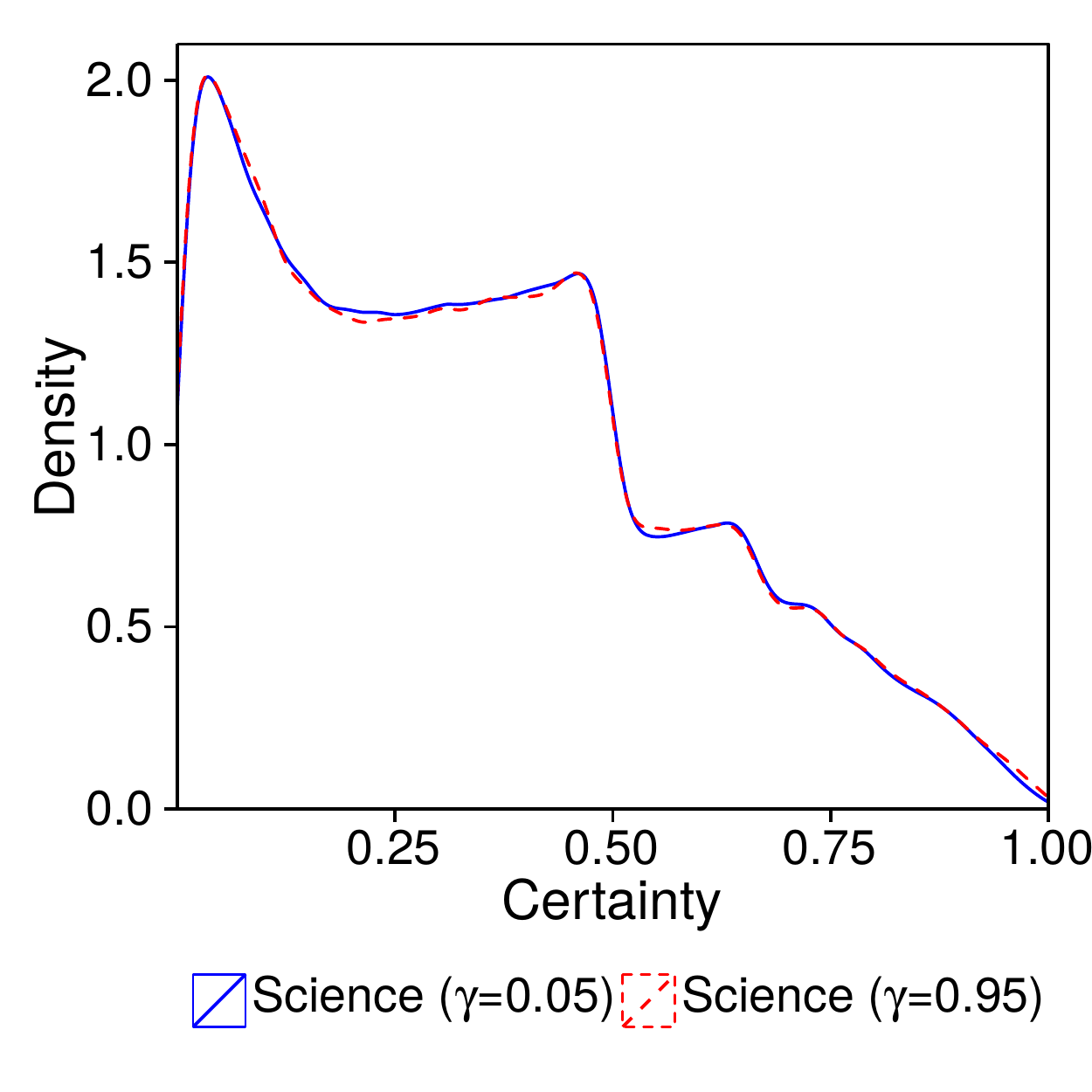}
	\hspace{1cm}
	\includegraphics[width=.40\textwidth]{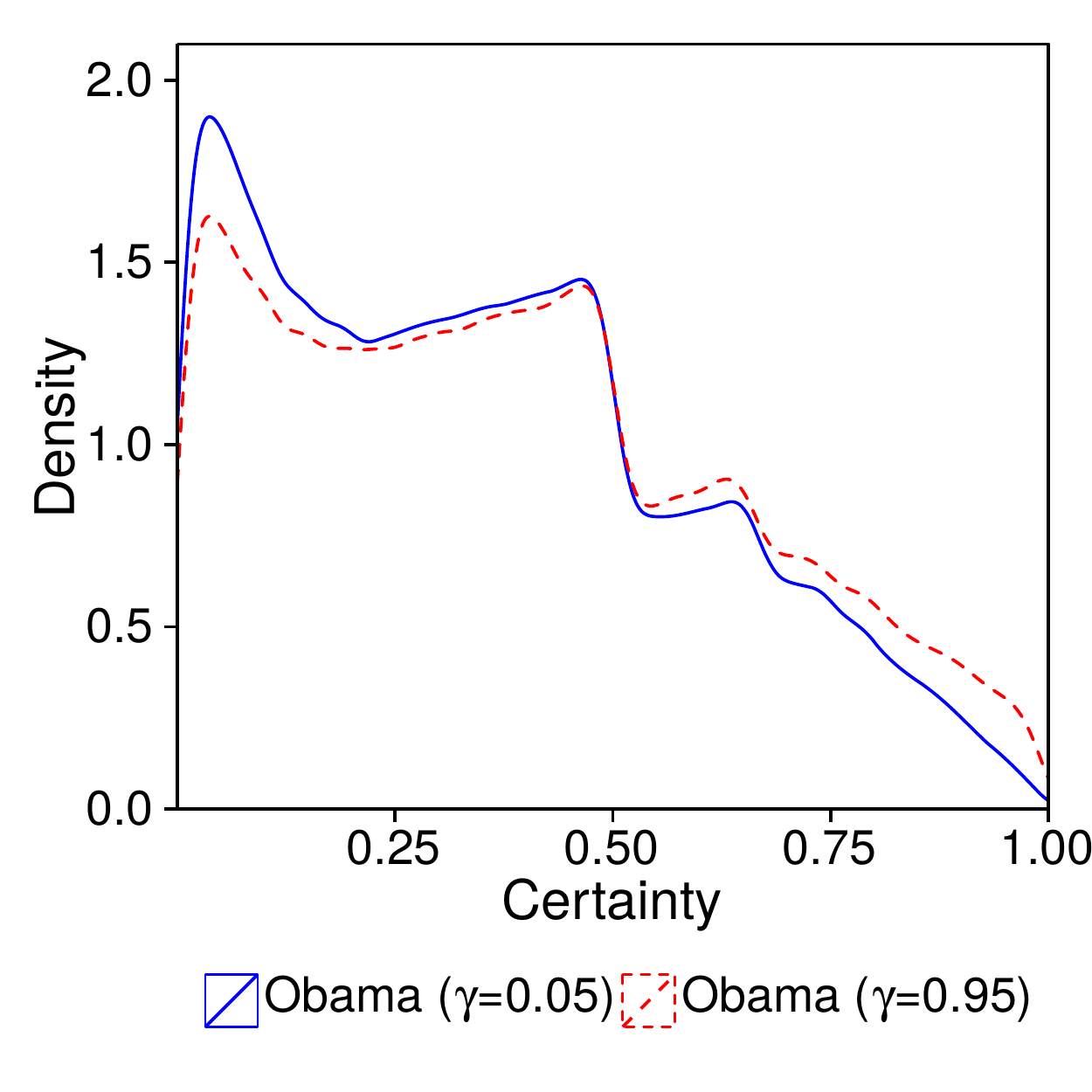}
	\caption{Probability density function of parent sample certainty.}
	\label{fig:wiki_certainty_density}
\end{figure}

Figure~\ref{fig:wiki_certainty_density} shows the probability density functions for the certainties in HDTMs parent sampling process. In the figures on both left and right we find that the probability densities appear to be polynomial distributions with interesting plateaus that end just before the 50\%, 66\%, and 75\% certainty scores corresponding to the near-certain (but not perfectly-certain) scores for nodes with indegree values of 2, 3, and 4 respectively. In these results, and others not displayed in this paper, we find that when the root is a general document like \textsf{Science}, then changing $\gamma$ does not affect the certainty distribution. Given a relatively specific root like \textsf{Barack Obama}, larger $\gamma$ values increase certainty overall.

As discussed earlier, measuring topical similarity in hierarchies can be precarious because, in many cases, documents that are correctly situated under a topically unrelated parent may still have a strong contextual association with the parent document that outweighs the topical/language-oriented similarity as determined by the $\gamma$ parameter. As an example, consider the Wiki-articles \textsf{Honest Leadership and Open Government Act} and \textsf{Alexi Giannoulias}: even though these two articles are topically dissimilar, HDTM parameterized with a high $\gamma$ value is likely to place both articles as children of the hierarchy's root \textsf{Barack Obama} because the high $\gamma$ values weigh the topological link as more important that the documents' inferred topicality. If, on the other hand, HDTM was parameterized with a very low $\gamma$ value, then \textsf{Alexi Giannoulias} is more likely to be situated with other state senators, and the \textsf{Honest Leadership and Open Government Act} is more likely to be situated with other legislation.

To quantitatively evaluate the topical and topological fitness of hierarchy together, we compare the similarily among the sets of parents and children in the inferred hierarchy against sets of Wiki-articles and their Wikipedia categories. Because we are comparing sets of items we use the Jaccard coefficient for this task. Specifically, the Jaccard coefficient is calculated by

\begin{equation}
 J_{coefficient} = \frac{|C_{d}\cap C_{pa}|}{|C_{d}|+|C_{pa}| - |C_{d}\cap C_{pa}|}, 
\end{equation}

for each document $d$ that belongs to a set of categories $C_d$, and where $C_{pa}$ is the union of the corresponding categories of the nodes in $d$'s ancestors $\mathbf{c}_d$. The Jaccard coefficient is useful here because it gives an quantitative measure describing how well the inferred hierarchical structure matches with the human-annotated categorical structure of Wikipedia. 

\begin{figure}
    \centering
    \includegraphics[width=.40\textwidth]{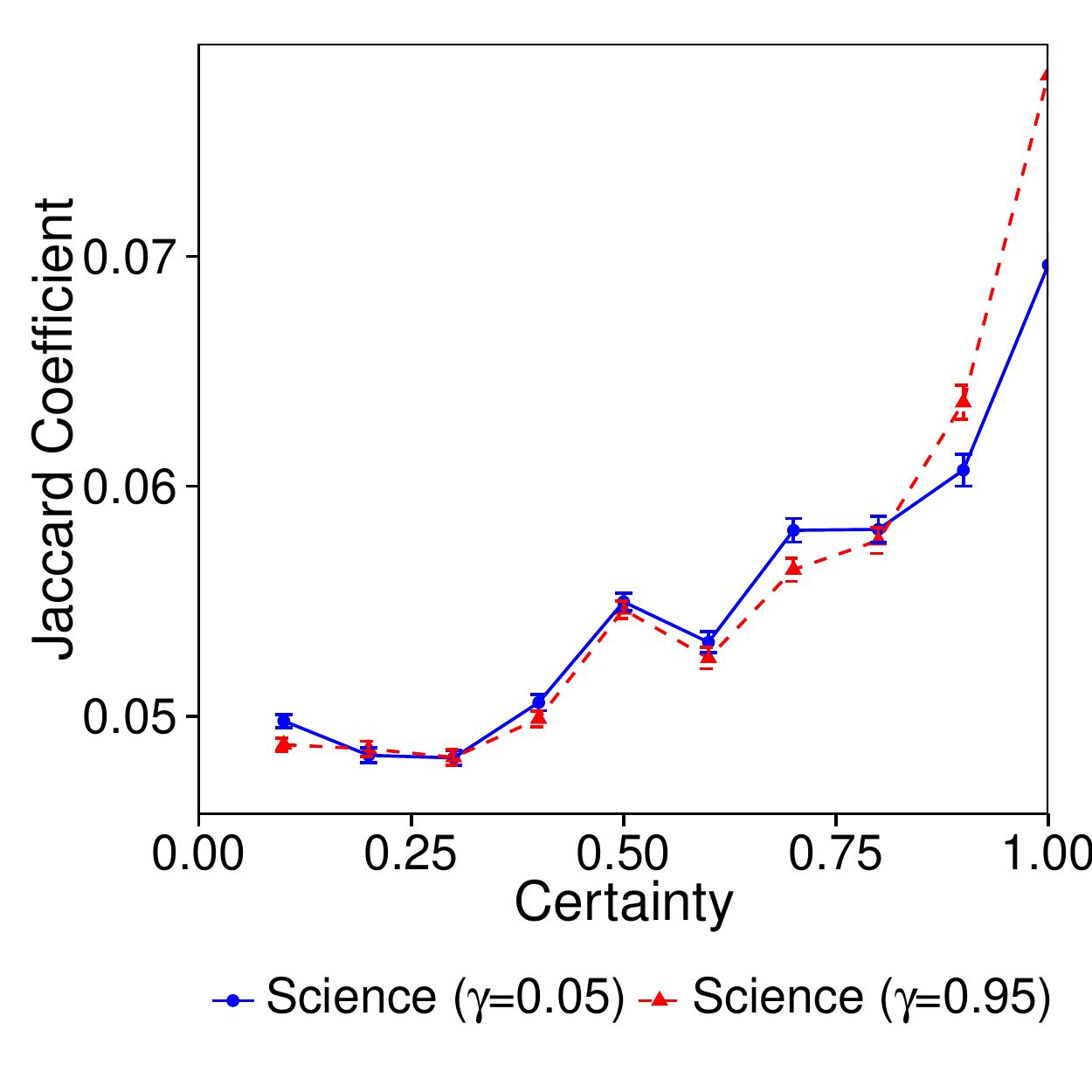}
	\hspace{1cm}
    \includegraphics[width=.40\textwidth]{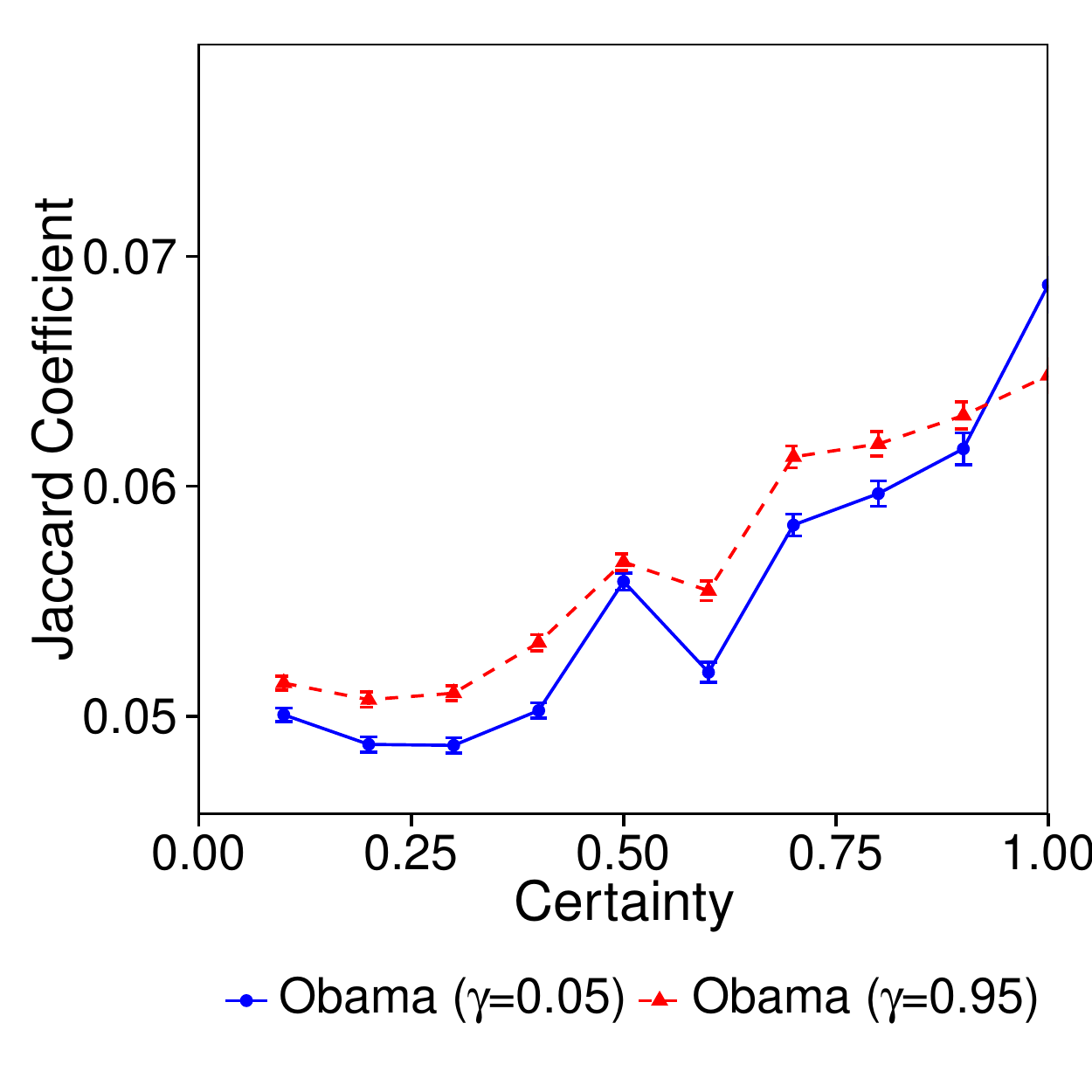}
    \caption{Jaccard coefficient versus certainty in 10 bins. Error bars are drawn for each bin, but many are too small to be visible. Higher score are better.}
    \label{fig:wiki_jaccard}
\end{figure}

Figure~\ref{fig:wiki_jaccard} shows the relationship between Jaccard coefficients of the hierarchies. For illustration clarity, in this figure all Jaccard coefficients are binned into 10 equal intervals. We observe a clear correlation between the node's certainty function and its Jaccard coefficient. This means that decisions that HDTM is most certain about tend to be the most similar to the human-created hierarchies.

Recall that HDTM uses the $\gamma$ parameter to balance the weight of topological links and topical similarities. Hence when specific nodes like \textsf{Barack Obama} are chosen as the root, higher $\gamma$ will reduce Jaccard coefficient for high certainty nodes. This is because higher $\gamma$ values favor topology rather than content during the path and parent selection process. 

These results also underscore the importance of the selection of the root node. When provided a root node, which can be any node in the document graph, the inferred hierarchy is created from the perspective of that root. The hierarchy rooted at \textsf{Barack Obama} would therefore organize, say, congressional legislation differently than a hierarchy generated by the Wiki-article on \textsf{Science}. Yet, as we've seen, the graph similarity metrics views the drastically different document-topic hierarchies as topologically similar.

\subsection{Discussion}
\label{subsec:discussion}
In order to properly understand the results captured in Table~\ref{tab:LLResults} from above, recall that log likelihood is a metric on the {\em fit} of the observations on the configuration of the model. The original work on LDA~\citep{Blei2003} found that likelihood increases as the number of topics increases. Along those lines, Chang, {\em et al.} demonstrated that more fine grained topics, which appear in models with a larger number of topics have a lower interpretability, despite having higher likelihood scores~\citep{Chang2009}. Simply put, there exists a negative correlation between likelihood scores and human interpretability in LDA and similar topic models.

Applying those lessons to our experiments, recall that HDTM has as many topics as there are documents, and non-root document topics are mixtures of the topics on the path to the root. Also recall that HLDA, TopicBlock and TSSB all generate a large number of latent topics. In HLDA and TopicBlock, there are infinitely many topics/tables in the nCRP; and practically speaking, the number of topics in the final model is much larger than the number of documents (conditioned on the $\gamma$ parameter). In TSSB, the topic generation is said to be an interleaving of two stick breaking processes; practically, this generates even larger topic hierarchies. The fsLDA algorithm has as many topics as there are in hLDA, however, the fsLDA hierarchy is not redrawn during Gibbs iterations to fit the word distributions resulting in a lower likelihood. Simply put, the number of topics in HDTM and fsLDA $=|V|$  $\ll$ hPAM, hLDA and TopicBlock $\ll$ TSSB.

Similarly, Figure~\ref{fig:llavgdepth} shows that deeper hierarchies have higher likelihood scores. This is because long document-to-root paths, found in deep hierarchies, are able to provide a more fine grained fit for the words in the document resulting in a higher likelihood.

Therefore, the better likelihood values of HLDA, TopicBlock and TSSB are due to the larger number of topics that these models infer and not necessarily due to better model outputs. In this case, a better way to evaluate model accuracy is by some external task or by manually judging the coherence of the topics.

\subsection{Qualitative Analysis}
To measure the coherence of the inferred groupings, the {\em word intrusion} task developed by Chang {\em et al}~\cite{Chang2009} is slightly modified to create the {\em document intrusion} task. In this task, a human subject is presented with a randomly ordered set of eight document titles. The task for the human judge is to find the intruder, that is, the judge is asked to find which document is out of place or does not belong. If the set of documents without the intruder document all make sense together, then the human judge should easily be able to find the intruder. For example, given a set of computer science documents with titles \{\textsf{systems, networking, databases, graphics, Alan Turing}\}, most people, even non-computer scientists, would pick \textsf{Alan Turing} as the intruder because the remaining words make sense together -- they are all computer science disciplines.

For the set \{\textsf{systems, networking, RAM, Minesweeper, Alan Turing}\}, identifying a single intruder is more difficult. Human judges, when forced to make a choice, will choose an intruder at random, indicating that the grouping has poor coherence.

\begin{figure}
	\centering
		\includegraphics[width=.95\textwidth]{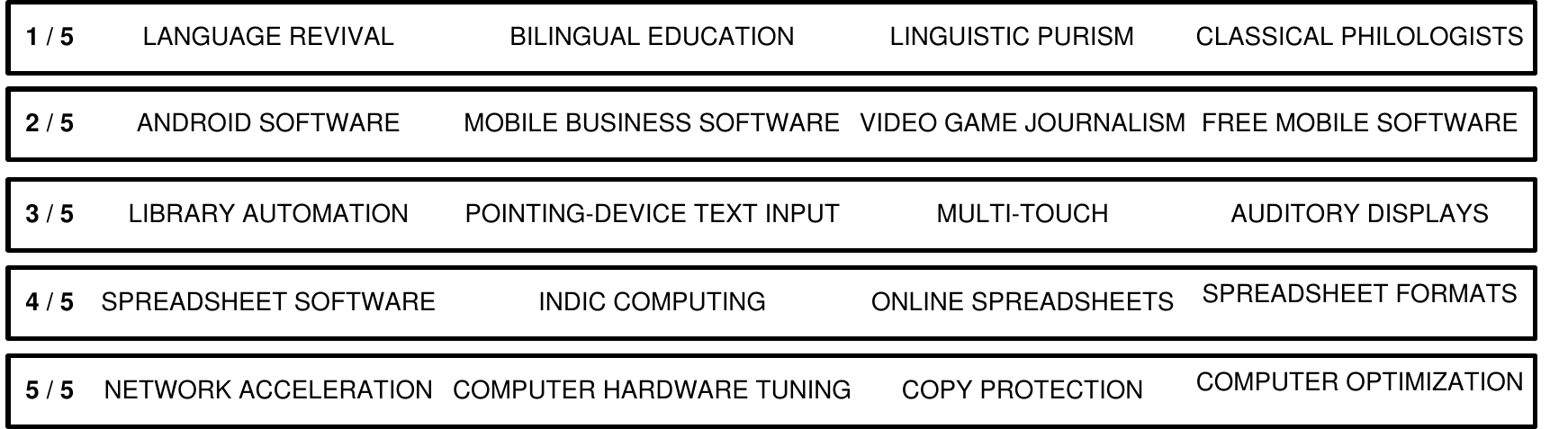}		
	\caption{Illustration of the intruder detection task from the Wikipedia collection, wherein human judges are presented with a set of document titles and asked to select the document that does not belong.}
	\label{fig:intruder}
\end{figure}

\begin{figure}
	\centering
		\subfigure[CompSci Web site]{
		\includegraphics[width=.65\textwidth]{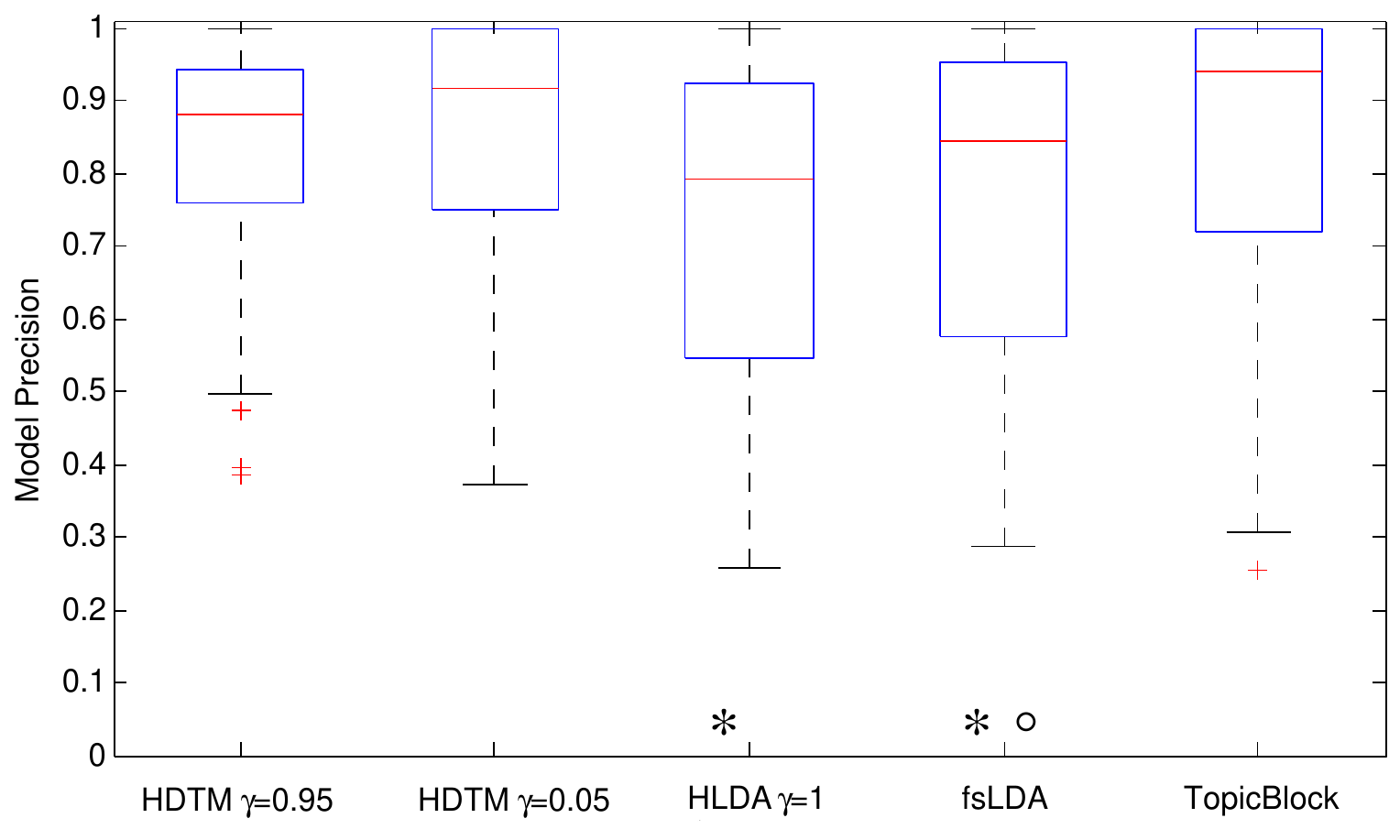}
		\label{fig:csturk}
		}
		\subfigure[Wikipedia]{
		\includegraphics[width=.65\textwidth]{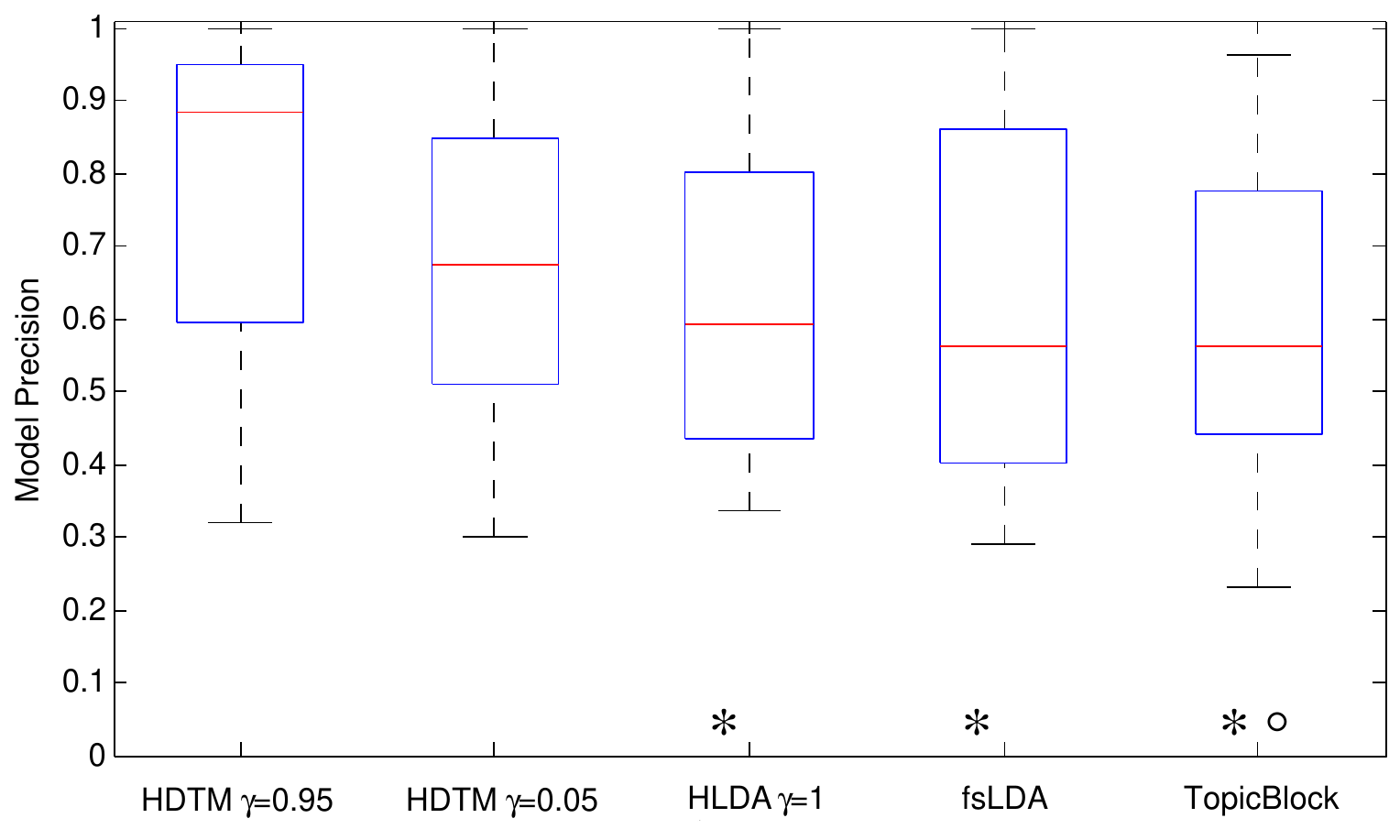}
		\label{fig:wikiturk}
		}
		\subfigure[Bib. Network]{
		\includegraphics[width=.65\textwidth]{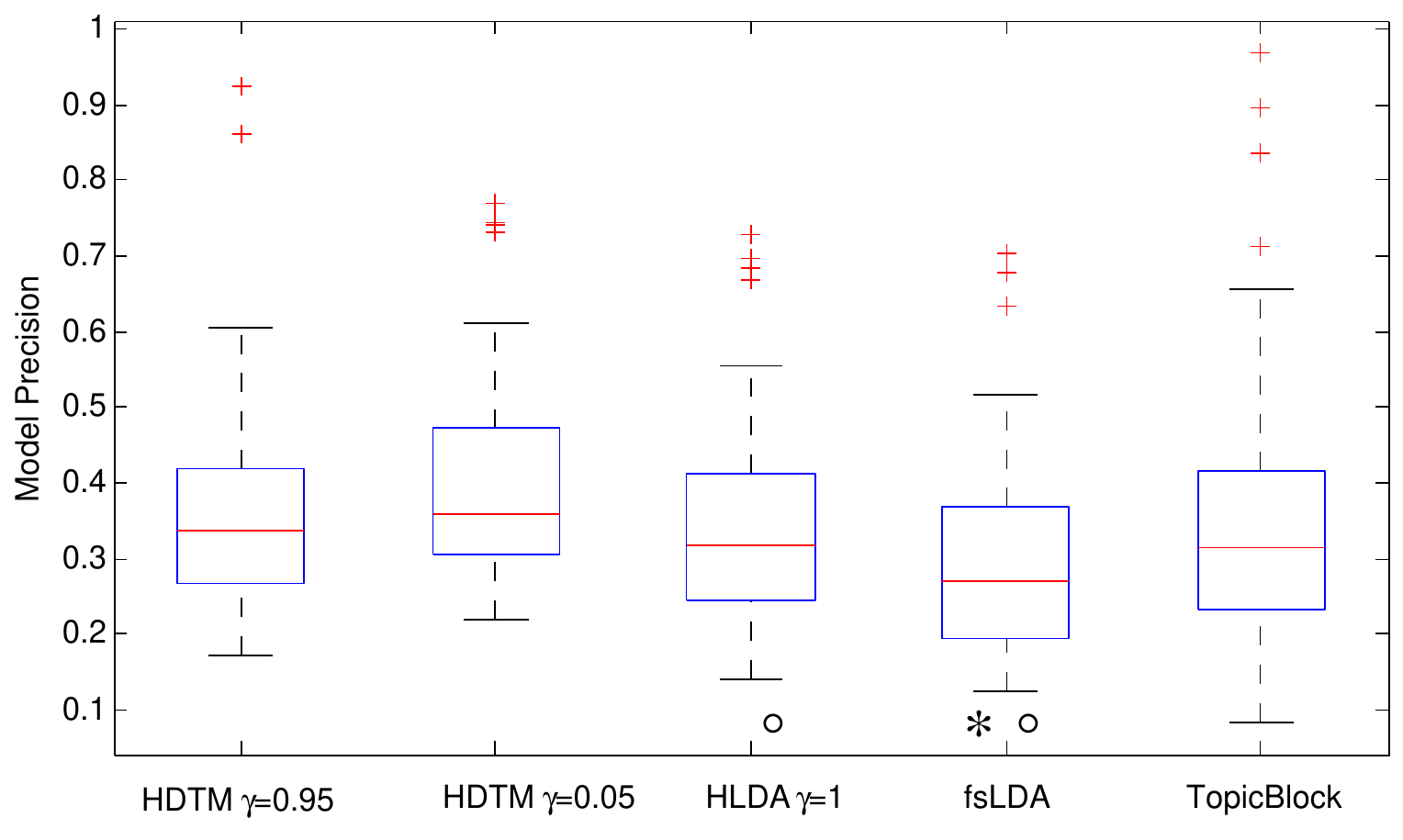}
		\label{fig:dblpturk}
		}
	\caption{The model precision for five models on three document-graph collections. Higher is better. $\ast$ and $\circ$ represents statistical significance from HDTM $\gamma = 0.95$ and $\gamma = 0.05$ respectively.}
	\label{fig:turks}
\end{figure}

To construct a set of document titles to present to the human judge, a grouping from the hierarchy is selected at random, and select seven documents are further selected at random from the grouping. If the there are fewer than 7 documents available in the selected grouping, then we select all of the documents available; groupings of size less than 4 are thrown out. In addition to these documents, an intruder document is selected at random from among the entire collection of documents minus the documents in the test group. Titles are then shuffled and presented to the human judges.

\subsubsection{Comparison Models}
\label{sec:grouping}
Extracting document groupings for evaluation is slightly different for each model. HDTM and fsLDA store a document at each node in the hierarchy. A grouping is selected by first picking a document at random, and then choosing its siblings. TopicBlock and HLDA store documents at the leaves of the taxonomy, which often include several documents. A grouping is selected from these models by first picking a document at random, and then choosing the other documents in the leaf-topic.

The hierarchies that the TSSB model constructed allowed multiple documents to live at inner nodes. Attempts to evaluate groupings on inner nodes with more than 4 documents were unsuccessful. Nodes with 4 or more siblings were also difficult to find because the hierarchies that were generated were too sparse to find practical groupings. Thus human judges with TSSB groupings could not be found.

Each document-graph collection had different types of labels presented to the judges. The CompSci web site collection was labeled by the Web Page title and URL; the Wikipedia collection was labeled by the category title as shown in Figure~\ref{fig:intruder}; the bibliography network was labeled by the title of the paper.

\subsubsection{Analyzing human judgments}

The intruder detection tasks described above were offered on Amazon Mechanical Turk. No specialized training is expected of the judges. 50 tasks were created for each data set and model combination; each user was presented with 5 tasks at a time at a cost of \$0.07 per task. Each task was evaluated by 15 separate judges. In order to measure the trustworthiness of a judge, 5 easy tasks were selected, {\em i.e.}, groupings with clear intruders, and gold-standard answers were created. Judges who did not answer 80\% of the gold-standard answers correctly are thrown out and not paid. In total the solicitation attracted 31,494 judgments, across 14 models of 50 tasks each. Of these, 13,165 judgments were found to be from trustworthy judges.

The {\em model precision} is measured based on how well the intruders were detected by the judges. Specifically, if the intruder word $w^m_k$ is from model $m$ and task $k$, and $i^m_{k,j}$ is the intruder selected by the human judge $j$ on task $k$ in model $m$ then 

\begin{equation}
    \textrm{MP}^m_k = \sum_J{\mathbbm{1}(i^m_{k,j} = w^m_{k})/J},
\end{equation}

\noindent where $\mathbbm{1}(\cdot)$ is the indicator function and $J$ is the number of judges. The model precision is basically the fraction of judges agreeing with the model.

Figure~\ref{fig:turks} shows boxplots of the precision for the four models on three corpora. In most cases, HDTM performs the best. As in~\cite{Chang2009}, the likelihood scores do not necessarily correspond to human judgments. Paired, two-tailed t-tests of statistical significants ($p<0.05$) performed between HDTM $\gamma=0.95$ and $\gamma=0.05$ and the other models are represented by $\ast$ and $\circ$ in Figure~\ref{fig:turks} respectively.

\begin{figure}
	\centering
		\includegraphics[width=.85\textwidth]{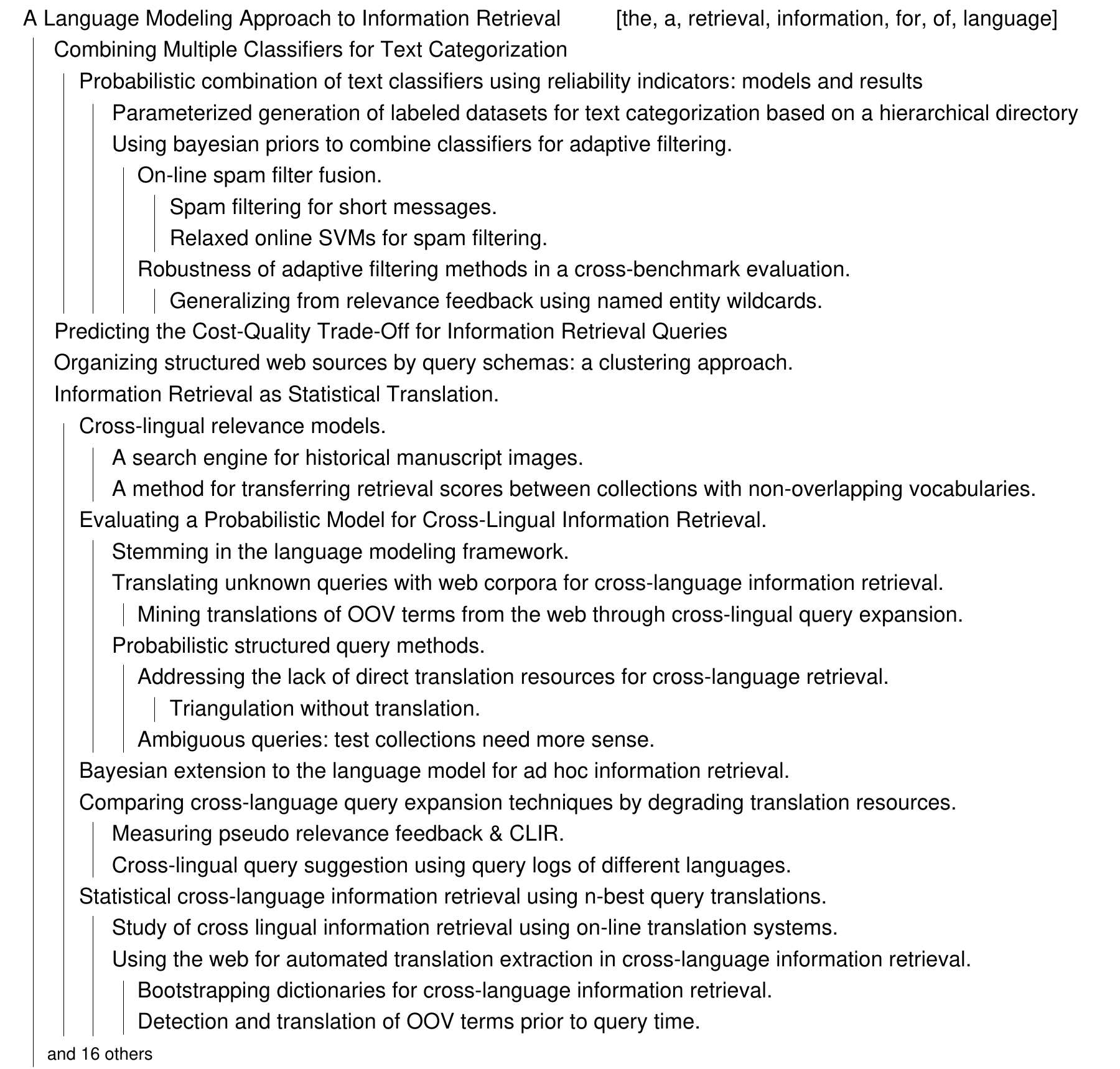}		
	\caption{Constructed hierarchy of bibliographic network with HDTM $\gamma=.95$. Words at the root document represent the most probable words in the root topic. Most probable words for other documents are not shown due to space constraints.}
	\label{fig:dblphierarchy}
\end{figure}

The bibliography network data had relatively low precision scores. This is almost certainly because it was more difficult for the judges, who were probably not computer scientists, to differentiate between the topics in research paper titles. Figure~\ref{fig:dblphierarchy} shows a small portion of the document hierarchy for the bibliographic network data set constructed with HDTM $\gamma=.95$. The root document has 20 children in the hierarchy despite having 145 in-collection links. The remaining 120 documents live deeper in the hierarchy because HDTM has determined that they are too specific to warrant a first level position, and have a better fit in one of the subtrees. 

Recall that each document is associated with the topics from itself to the root, where the root is a single, general topic. The seven most probable terms at the root level are also shown adjacent to the root's title in Figure~\ref{fig:dblphierarchy}. These terms, like in HLDA and TopicBlock, are terms that are general to the entire collection. Similar sets of words exist at each node in the hierarchy, but are not shown in this illustration to maintain clarity.

\subsection{Reproducibility}
HDTM source code, analysis code and the scripts which generated the results found in this paper can be downloaded from {\small \url{https://github.com/nddsg/HDTM/releases/tag/kais}}. The Wikipedia, Web site and bibliographical data are all publicly available and free to download and are replicated in our own data repository.

\section{Conclusions}
\label{sec:conclusions}
Hierarchical document-topic model (HDTM), is a Bayesian generative model that creates document and topic hierarchies from rooted document graphs. The initial hypothesis was that document graphs, such as Web sites, Wikipedia and bibliographic networks contain a hidden hierarchy.  Unlike most previous work, HDTM allows documents to live at non-leaf nodes in the hierarchy, which requires the random walk with restart path sampling technique. An interesting side-effect of the random walker adaptation is that the path sampling step, Eq.~\ref{eq:pathsample}, is much faster and easier to scale than the nCRP because RWR only creates a sampling distribution for the parents of a document, whereas the nCRP process creates a sampling distribution over all possible paths in the taxonomy.

Several quantitative experiments comparing HDTM with related models were performed. However, the results show that likelihood scores are a poor indicator of hierarchy interpretability, especially when the number of topics are different between comparison models. A large qualitative case study was also performed which showed that the cohesiveness of the document groupings generated by HDTM were statistically better than many of the comparison models despite the poor likelihood scores.

\begin{acknowledgements}
This work is sponsored by an AFOSR grant FA9550-15-1-0003, and a John Templeton Foundation grant FP053369-M.
\end{acknowledgements}

\end{document}